\documentclass[a4paper,12pt,twoside]{svmult}
\newcommand{\bea}{\begin{eqnarray}}
\newcommand{\eea}{\end{eqnarray}}
\newcommand{\be}{\begin{equation}}
\newcommand{\ee}{\end{equation}}
\newcommand{\bd}{\begin{displaymath}}
\newcommand{\ed}{\end{displaymath}}
\newcommand{\bi}{\begin{itemize}}
\newcommand{\ei}{\end{itemize}}
\newcommand{\ba}{\begin{array}}
\newcommand{\ea}{\end{array}}

\newcommand{\bbox}{\begin{tcolorbox}}
\newcommand{\ebox}{\end{tcolorbox}}

\usepackage[left=2cm,right=2cm,top=2cm,bottom=3cm]{geometry}
\usepackage{tcolorbox}
\usepackage{tikz}
\usepackage[english]{babel}
\usepackage{fancyhdr}
\usepackage{titlepic}
\usepackage{fancyhdr}
\usepackage{graphicx}
\usepackage{lipsum}
\usepackage{titling}
\usepackage{textcomp}
\usepackage{gensymb}
\usepackage{amsfonts}
\usepackage{multirow}
\usepackage{subcaption}
\usepackage{siunitx}
\usepackage{microtype}  
\usepackage[intoc]{nomencl}
\usepackage{color}
\input{ImpStyle.sty}
\usepackage{keystroke}
\usepackage{listings}
\usepackage{setspace}
\usepackage{amsmath}
\usepackage{upgreek}
\usepackage{lscape}
\usepackage{pbox}
\usepackage[tableposition=top]{caption}
\usepackage{float}
\floatstyle{plaintop}
\restylefloat{table}
\usepackage[version=3]{mhchem}
\usepackage[sectionbib]{chapterbib}
\usepackage{algorithm}
\usepackage[hidelinks]{hyperref}
\hypersetup{
    colorlinks=false, 
    linktoc=all,     
    linkcolor=black,  
}

\makeatletter
\newcommand{\chapterauthor}[1]{%
  {\parindent0pt\vspace*{-25pt}%
  \linespread{1.1}\large\scshape#1%
  \par\nobreak\vspace*{35pt}}
  \@afterheading%
}
\makeatother
\usepackage{afterpage}
\usepackage{booktabs}
\usepackage{listings}
\usepackage{hyperref}
\usepackage{makecell}
\usepackage{threeparttable}
\usepackage{algpseudocode}
\usepackage{mathtools}
\usepackage{amsmath}
\usepackage{amssymb}
\usepackage{physics}
\usepackage{wasysym}
\usepackage{array}
\usepackage{tabularx}
\algnewcommand\algorithmicforeach{\textbf{for each}}
\algdef{S}[FOR]{ForEach}[1]{\algorithmicforeach\ #1\ \algorithmicdo}

\DeclareRobustCommand{\ppp}[0]{%
\begin{tikzpicture}[line width=0.3pt, scale=1.2]%
\draw (0ex,0ex) -- (1.5ex,0ex);
\draw (1.5ex,0ex) -- (0.75ex,1.5ex);
\draw (0.75ex,1.5ex) -- (0ex,0ex);
\end{tikzpicture}}

\DeclareRobustCommand{\pppp}[0]{%
\begin{tikzpicture}[line width=0.3pt, scale=1.2]%
\draw (0ex,0ex) -- (1.5ex,0ex);
\draw (1.5ex,0ex) -- (0.75ex,1.5ex);
\draw (0.75ex,1.5ex) -- (0ex,0ex);
\draw (0.75ex,0.65ex) -- (0ex,0ex);
\draw (0.75ex,0.65ex) -- (1.5ex,0ex);
\draw (0.75ex,0.65ex) -- (0.75ex,1.5ex);
\end{tikzpicture}}

\newcommand\myeq{\stackrel{\mathclap{\scriptsize\mbox{def}}}{=}}

\usepackage{atbegshi}
\AtBeginDocument{\AtBeginShipoutNext{\AtBeginShipoutDiscard}}
\title{\LARGE {Handbook of Soft Robotics}\\
 \vspace*{6mm}}

\author{\\Thrishantha Nanayakkara,~ Jamie Paik,~ Barbara Mazzolai,~ Cecilia Laschi,~\\Manu Srivastava,~Ian Walker,~  Shinichi Hirai,~ \\Christian Duriez,~ Isuru Godage,~ Hunter Gilbert,~\\Nicolas Rojas,~ Hadi Sadati,~ Ali Starbanov }
\usepackage{tikz}

\begin{document}






\setcounter{chapter}{6}
\setcounter{page}{168}
\chapter[Malleable Robots]{Malleable Robots}\label{chMalleableRobots}
\chapterauthor{Angus B. Clark, Xinran Wang, Alex Ranne, Nicolas Rojas}

\textbf{Abstract } This chapter is about the fundamentals of fabrication, control, and human-robot interaction of a new type of collaborative robotic manipulators, called malleable robots, which are based on adjustable architectures of varying stiffness for achieving high dexterity with lower mobility arms. Collaborative robots, or cobots, commonly integrate six or more degrees of freedom (DOF) in a serial arm in order to allow positioning in constrained spaces and adaptability across tasks. Increasing the dexterity of robotic arms has been indeed traditionally accomplished by increasing the number of degrees of freedom of the system; however, once a robotic task has been established (e.g., a pick-and-place operation), the motion of the end-effector can be normally achieved using less than 6-DOF (i.e., lower mobility). The aim of malleable robots is to close the technological gap that separates current cobots from achieving flexible, accessible manufacturing automation with a reduced number of actuators.

\section{Introduction}\label{secIntro}
Reconfigurable robot systems provide several key potential advantages over traditional robots, including increased task versatility by adapting to better suit tasks, and reduced robot cost due to a smaller total number of modules, such as links and joints. As such, there has been significant research into the development of reconfigurable robots, with the most popular approach utilising modularity as the method of reconfiguration, as this allows for the interchangeability of parts, leading to self-repair \cite{yim2007modular, seo2019modular}. The reconfigurability feature has specifically been of interest in unstructured and unpredictable environments, characterised by changing operating contexts, which take the most advantage from robots that can adapt their shape and operating mode \cite{valente2016reconfigurable}.

An alternative approach for the application of reconfigurable robot manipulators can be found in the industrial field of serial manipulators. In an ideal case, a manipulator would be designed with the exact number and configuration of joints necessary for its expected set of tasks \cite{kereluk2017task}. This is known as task-based optimisation, and requires information to be known about the robot structure \cite{brandstotter2018task}, collections of working points \cite{yang2000task}, or end effector regions \cite{kereluk2017task}. However, knowledge of all tasks a robot might encounter in its lifetime can be difficult to determine. Instead, serial manipulators with a higher degree of freedom (DOF) are typically selected, ensuring dexterity across tasks at the expense of an increased robot cost and footprint.

\begin{figure}[t!]
    \centering
    \includegraphics[width=\columnwidth]{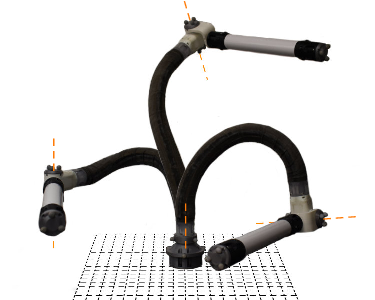}
    \caption{A two-degree-of-freedom (DOF) malleable robot arm, showing various topology configurations it can achieve. The positioning of the revolute axes for each configuration are highlighted.}
    \label{malleablerobotdiagram}
\end{figure}

Malleable robots, which are \textbf{reduced DOF serial robot arms with changeable geometry}, provide a solution to the task versatility problem through variable relative positioning of the revolute joints \cite{clark2020design}. An example of the reconfiguration capability of a 2-DOF malleable robot can be seen in Fig.~\ref{malleablerobotdiagram}. Malleable robots follow a similar process of reconfiguration to modular reconfigurable robots, in that their reconfiguration alters the relative positioning of the active joints of the robot. However, unlike modular reconfigurable robots which achieve this by disassembling and reassembling (in an alternative configuration) the modules of the robot, malleable robots achieve this by transitioning between reconfigurable and rigid modes. Once in a reconfigurable mode, the robot can either reconfigure itself using additional drive systems (intrinsic malleable robots), or it can be manually reconfigured by an external system, such as a user (extrinsic malleable robots) as discussed in \cite{clark2020design}, where the design of such robots was firstly explored.

This chapter will explore malleable robots, providing further details and discussion on their design and construction, their kinematics, the use of augmented reality to aid their reconfiguration, and their intrinsic control. For each section of this chapter a relevant tutorial is given, providing the reader with the necessary skills and information to become familiar with the full development process of malleable robots.

\section{Design of Malleable Robots}\label{MalleableDesign}
The key enabling technology behind malleable robots is their malleable link. This is defined broadly as a segment or link that connects two active revolute robot joints at an offset, ensuring their coordinate axes are different. The most important aspect of malleable robots is that this offset between their coordinate axes is \textbf{variable} in all 6 degrees of freedom (3 translation, 3 rotation). It can be changed to modify the joints relative positions, which in turn modifies the structure of the robot, resulting in a variable but controllable workspace, which can form specialised topologies (e.g., SCARA,  Spherical, or PUMA-like). In turn, this enables malleable robots to be of lower mobility than typical serial robot manipulators, where the majority are composed of rigid links and joints \cite{majidi2014soft}, as they are highly flexible and can adapt to different tasks.

It is also key that this reconfiguration only occurs when desired, and at all other times the structure of the robot is fixed. As described, there is no one specific method for creating a malleable link. Various technologies could be used to provide the variable structure of the link, examples including simple lockable multi-segment links \cite{aghili2009reconfigurable} and more complex variable stiffness technologies such as jamming particles, the latter of which we will examine more herein. Considering how they are reconfigured, malleable links can be classified into two categories: \textit{intrinsic}, where the reconfiguration of the link is performed by the link itself, such as through an internal actuation system, and \textit{extrinsic}, where the reconfiguration is performed by an external factor, such as a human manually reconfiguring the robot.

In this section, we will explore the alternative solutions in the design and construction of malleable robots, and provide a guide for an extrinsic layer jamming based solution.

\subsection{Variable Stiffness Technologies}
The promising performance of soft continuum robots has led to the development of various continuum implementations \cite{kang2013design, grissom2006design, mcmahan2005design}, however despite providing increased dexterity they suffer from a number of issues, most notably a lack in holding strength \cite{robertson2017soft, burgner2015continuum}. To solve this, soft continuum robots have integrated variable stiffness technologies within their design, allowing for both soft, highly flexible motion and rigid, high load resistance \cite{jiang2012design, santiago2016soft, kim2013novel, kim2012design}. These technologies are discussed and categorised in depth by L. Blanc et al. \cite{blanc2017flexible}.

Variable stiffness technologies provide high flexibility and high stiffness in typically a small space, and have been used to increase the stiffness of small flexible devices. As such they have shown significant development in the medical device sector for increasing the stiffness of medical endoscopes and tools when performing delicate tasks requiring a high force output (and thus a high stiffness) during surgery.

Malleable links are typically larger (>50mm diameter) than standard medical continuum robots (<15mm diameter). This has consequences on the variable stiffness technologies that can be used for malleable links, as their increase in size means that certain technologies, like shape memory alloys (SMAs) and low melting point alloys (LMPAs), are unsuitable due to the significant increase in time to transition between rigid and flexible states.

A comparison on different variable stiffness technologies, along with their considerations regarding malleable links, can be seen in Table~\ref{tab:variable_stiffness}

\begin{table}[!t]
    \small
    \centering
    \rowcolors{2}{gray!25}{white}
    \begin{tabular}{p{17mm}p{2cm}p{3cm}p{3cm}}
         \textbf{Method} & \textbf{Example Technology} & \textbf{Pros} & \textbf{Cons} \\
         \hline
         Glass Transition & SMA & Extensive existing literature and use. & Activation time increases significantly with scale. Expensive.\\
         Phase Transition & LMPA & Large stiffness gain. & Activation time increases significantly with scale. \\
         Rheological Fluids & Magnetorheological & Can achieve very fast (millisecond) activation times. & Challenging to seal and maintain for long term use.\\
         Bulk Locking & Granular Jamming & Very easy to assemble. Quick and simple to control stiffness. & Weight of the granules can become a challenge with significant increase in scale.\\
         Longitudinal Locking & Layer Jamming & Lightweight, hollow centre. & Complex to assemble.\\
         Segments Locking & Central Wire & Simple to actuate, as can be controlled by a single wire. & Backlash in the segments is cumulative, resulting in significant backlash for long links. \\
         \hline
    \end{tabular}
    \caption{Variable stiffness technologies}
    \label{tab:variable_stiffness}
\end{table}

Of these technologies, longitudinal locking is the most promising for use in malleable links, due to the limited disadvantages of significantly scaling it up. For malleable links, due to their use as a robot manipulator and possible extrinsic reconfiguration, we suggest their diameter is as existing manipulator links at 40-50mm, which are also the optimal diameter for manual manipulation \cite{mcdowell2012effects}. First proposed by Y-J. Kim et al. \cite{kim2012design}, layer jamming is a promising technology that utilises longitudinal locking of thin layers of material wrapped conically along a cylinder to achieve a significant stiffness gain in a very thin and lightweight structure. Thanks to these features, it has shown recent popularity in continuum robotics \cite{ yoon2009flexible, mehling2006minimally, blessing2004novel, kim2013stiffness, santiago2015continuum, langer2018stiffening}.

One of the limitations of longitudinal locking is the stiffness dependence on the structure configuration, such as layer orientation. For continuum structures, this presents as a reduced stiffness as the cylindrical design bends. At extreme curvatures, buckling and unwanted deformation of the structure occurs, resulting in a dramatic loss of structural integrity. Thankfully, through the inclusion of a support structure within the continuum link, the stiffness decrease can be limited, and the buckling can be prevented \cite{clark2019stiffness}. It is important that, for such a design, it does not significantly affect the advantages of the continuum structure (hollow and light), nor does it reduce the available DOF ability.

\subsection{Manufacturing Flexible Spines and Malleable Links}

In this section we detail a method for achieving a variable stiffness layer-jamming-based malleable link, and a method for achieving a multi-DOF neutral-line constant diameter support mechanism, namely a flexible spine, to reduce structural deformation.

\subsubsection{Malleable Links}
\begin{figure}[t]
    \centering
    \includegraphics[width=0.95\columnwidth]{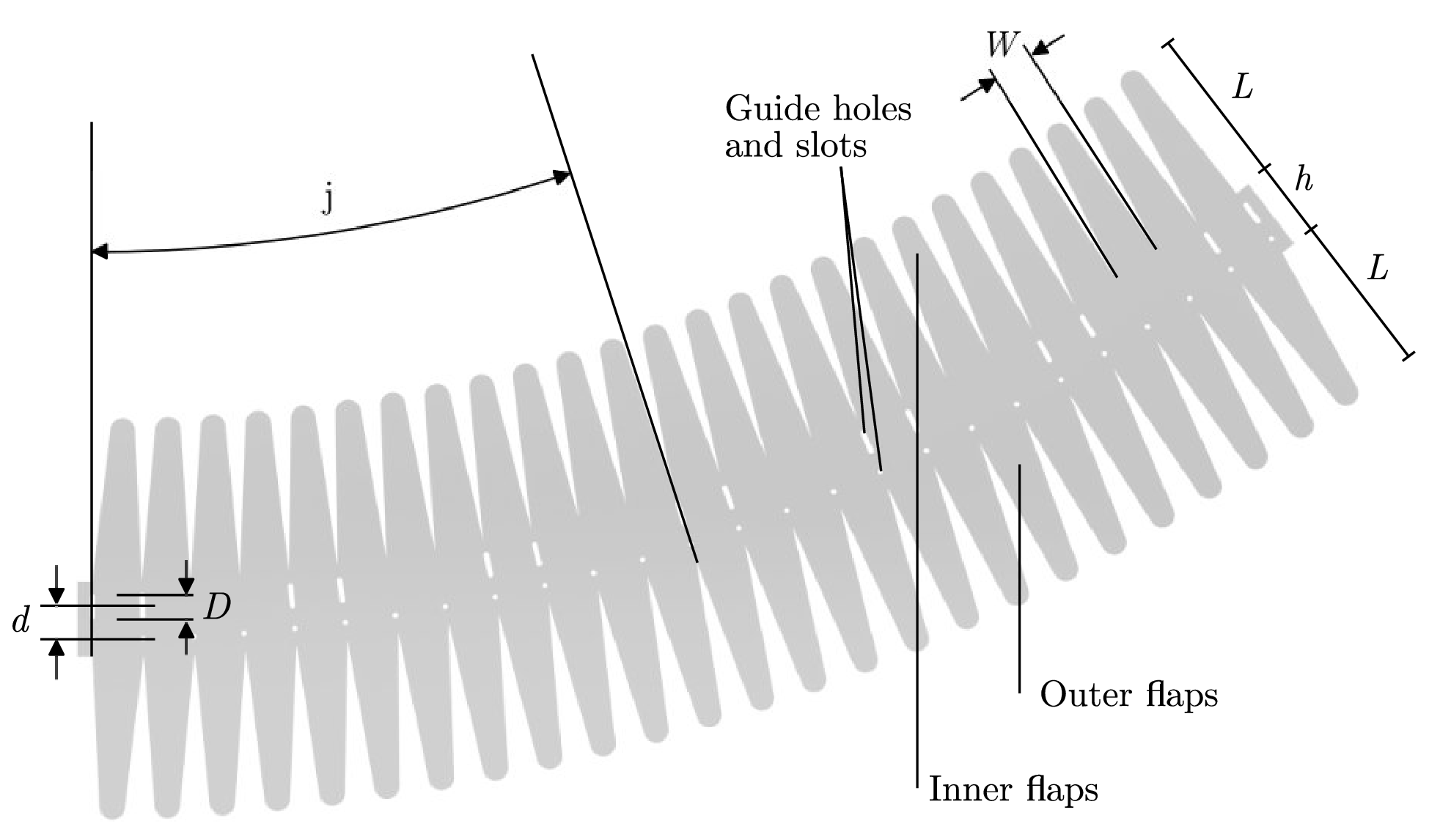}
    \caption{Double-sided flap pattern specifications for layer jamming sheath with guide holes and slots.}
    \label{LayerDesign}
\end{figure}

\begin{figure}[t!]
    \centering
    \includegraphics[width=0.9\columnwidth]{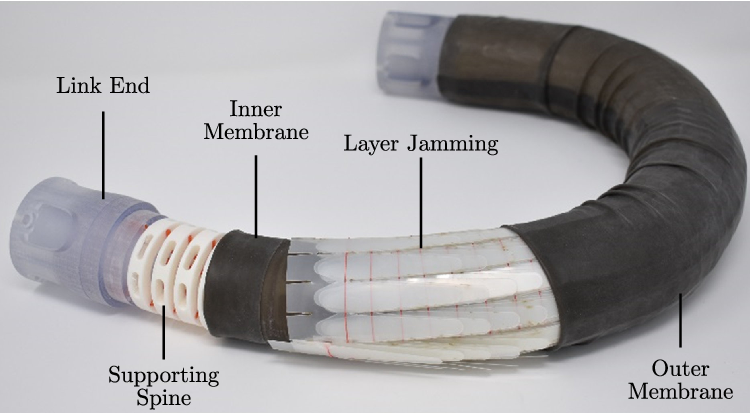}
    \caption{Cut-away malleable link highlighting individual components.}
    \label{malleablelink}
\end{figure}

The following approach to link design for extrinsic malleable robots focused on leveraging existing layer jamming technology used in continuum manipulators to create a finite length variable stiffness link, that is reconfigured using manual human deformation. Layer jamming using conically wrapped overlapping sheets of Mylar\textsuperscript{\textregistered} (0.18~mm) (polyethylene terephthalate), first presented as a medical manipulator by Kim et al. \cite{kim2012design} was used. The laser cut flap pattern, detailed in Fig.~\ref{LayerDesign}, contained 12 flaps spanning the circumference of the link, with a minimum of 10 overlapping layers always in contact. This determines the maximum stiffness, which can be calculated from the force $F$ required to separate the layers as $F=\mu nPWL$, where $\mu$ is the coefficient of friction, $n$ is the number of overlapping layers, and $P$ is the applied pressure. By varying the pressure in the sealed volume containing the layers using a vacuum pump (BACOENG 220V/50Hz BA-1 Standard), it is possible to vary the stiffness of the link from flexible to rigid as the cumulative friction causes a significant increase in rigidity proportional to the negative pressure applied. Flap parameters used were flap length $L=45mm$, flap width $W=13mm$, mid length $h=16mm$, guide hole distance $d=9.5mm$, and inclination angle $\varphi=12.75^\circ$. The flap pattern was then wrapped conically and contained within two cylinder membranes of latex sheet (0.25 mm), and sealed with link termination ends 3D printed from Vero Clear on a Stratasys Objet 500, which also provided mounting points for an internal structural spine (see next section) to prevent excessive deformation under extreme bending of the link, as well as mounting points to attach the other components of the robot. The components of the malleable link can be seen in Fig.~\ref{malleablelink}.

A key consideration of layer jamming is the termination of a finite length of it. While layer jamming provides a high level of stiffness when under compression, the point at which the layer jamming fixes to the rigid joints of the robot (which we call 'layer termination') presents a significant reduction in that stiffness. Due to the conical design, the termination point (and therefore fixture location) is not obvious. The traditional solution is to cut the tubular structure to form a flat face, which can then be adhered to. However, this results in a loss of $\sim$half of the overlapped layers at the termination point, resulting in a significant loss of stiffness. To correct this, passive layers can be used which replace the removed layers, but do not impact the termination.

The layers were manufactured by laser cutting matte surfaced polyethyleneterephthalate film (Mylar\textsuperscript{\textregistered}), with a thickness of 0.18mm and a measured coefficient of friction of 0.4. Extra strong polyester line (G\"utermann M 782) with a diameter of 0.20mm was sewn using a needle along the guide holes, completing a circumference every 12 holes. As the layers wrapped around forming a hollow cylinder, the guide slots of the previous loop were sewn to the following loop every 4 holes, forming connections every 120\textdegree. 

The slot length and guide hole distance $d$ determine the overall maximum and minimum length of the link through extension and compression, and as such the bending ability of the link (without deforming), where one side of the structure is at maximum extension while the other is at maximum compression. The maximum bending angle $\theta$ can be calculated as $\theta = (N-1)\sinh{\frac{D}{\phi}}$,
%
 where $N$ is the number of layer loops, $D$ is the slot length, and $\phi$ is the diameter of the layer jamming sheath. The maximum length $l_{max}$ and minimum length $l_{min}$ of the link can be computed as $l_{max},l_{min} = (N-1)(d\pm\frac{D}{2})+h$.
%
%

To modulate the stiffening mechanism, an outer and inner tubular membrane were joined at each end around the layers, forming a sealed tubular volume which could be pressurised, compressing the layers together. The membranes were manufactured manually by cutting sheet latex (0.25mm thickness) to the required length and perimeter, then sealing the latex into tubular membranes using liquid latex adhesive (CopyDex). At each end of the link, the layers and membranes terminated at a link end. These link ends were fabricated by 3D printing---fused deposition modeling (FDM) using ABS material, and incorporated a mounting position for both the membranes and the layers, which were attached using Ethyl 2-cyanoacrylate (Super Glue). Further, one of the two link ends enabled a connection via a 6mm PVC tubing to a vacuum pump, allowing access to the sealed volume containing the layers.

\subsubsection{Flexible Spine}

\begin{figure}[!t]
    \centering
    \includegraphics[width=0.95\columnwidth]{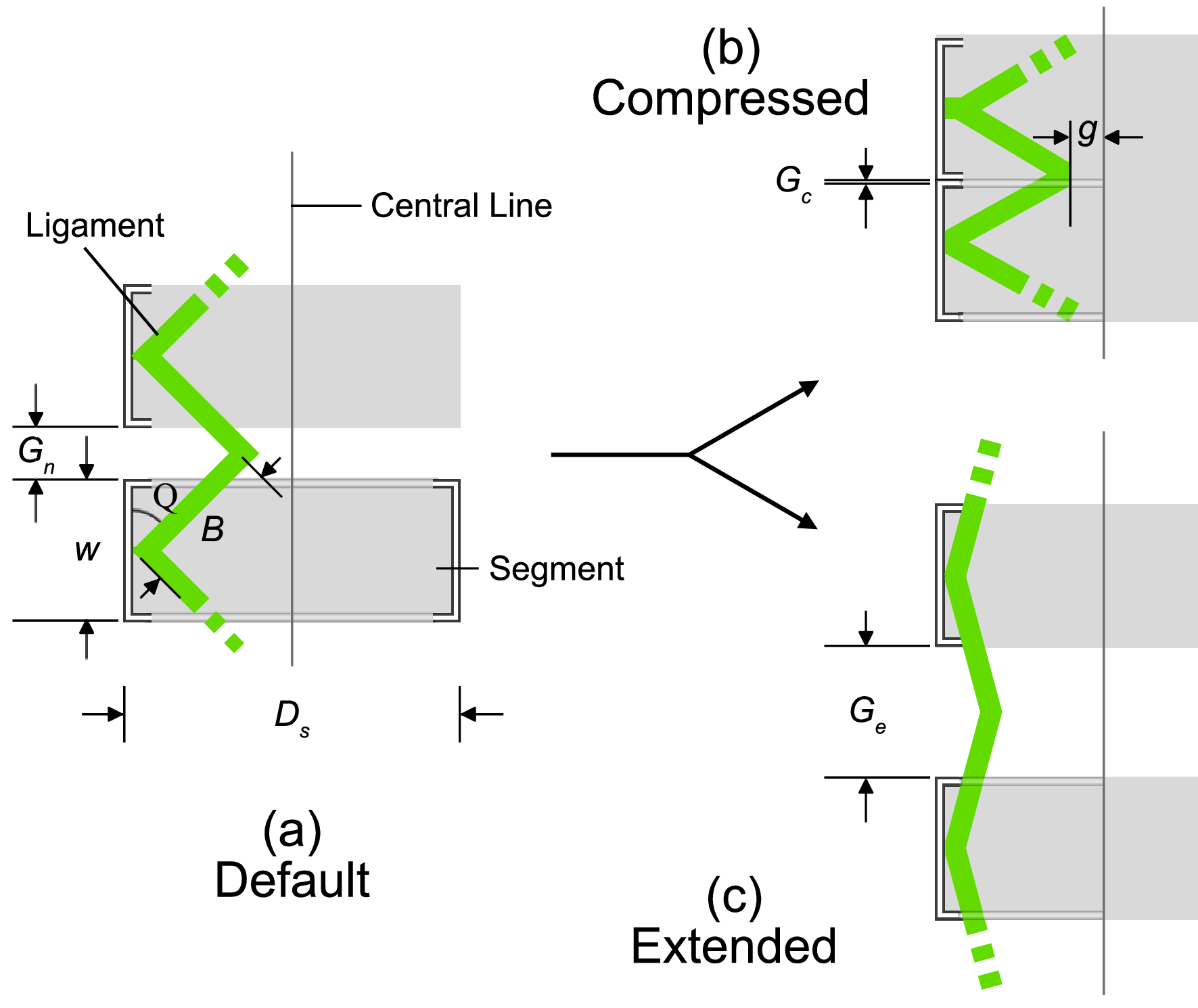}
    \caption{Neutral spine position {\bf (a)}, compressed spine {\bf (b)}, and extended spine {\bf (c)}.}
    \label{SpineDesign}
\end{figure}

To prevent buckling of the structure from the unmanaged deformation of the layers when bending, a supporting flexible spine was fabricated, which connected to each of the link ends, and maintained the diameter along the link as it underwent bending. The spine properties were identified from the layer jamming tube, making sure that the spine did not impact the existing bending performance of the malleable link. The spine is defined by the parameters neutral gap ($G_n$), compressed gap ($G_c$), extended gap ($G_e$), ligament beam length ($B$), ligament neutral angle ($\Theta$), central gap ($g$), segment height ($w$), and segment diameter ($D_s$), as shown in Fig.~\ref{SpineDesign}. To ensure the spine DOF did not affect the layers, an equal or lower rigid:flexible component ratio was required, to allow the spine equal or greater compression and extension than the layers. This was possible by using the maximum and minimum link lengths ($l_{max}$, $l_{min}$). This was computed assuming the flexible components were capable of compressing to a length of zero, and extending to the required length of $l_{max}-l_{min}$, implying an increase of 100\% from rest position. The spine was constructed from 14 rigid ABS segments, each 8mm in height and 32.5mm in diameter, totalling to 132mm of rigid support when added to the 10mm of support provided by each link end.

\begin{figure}[!t]
    \centering
    \includegraphics[width=\columnwidth]{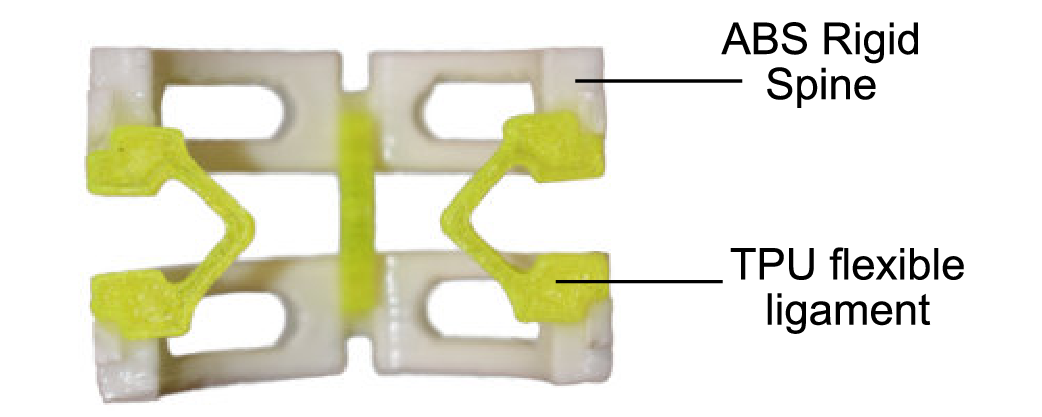}
    \caption{Sectional view of 2-segment spine with integrated flexible ligaments (square planar ligament connections)}
    \label{SpineCross}
\end{figure}

Connecting the segments were 3 magazine spring-inspired TPU ligaments, which were fixed every 120\textdegree\ to the segments and link ends using Ethyl 2-cyanoacrylate (Super Glue). The assembly of the components is demonstrated in Fig.~\ref{SpineCross}. The flexible ligaments were composed of connectors and triangular springs. When fully compressed, the central gap $g$ can be computed as
\begin{equation}
    \label{central-gap}
    g = \frac{D_s}{2}-(B^2-(G_c+\frac{w}{2})^2).
\end{equation}

\begin{figure}[t!]
    \centering
    \includegraphics[width=0.8\columnwidth]{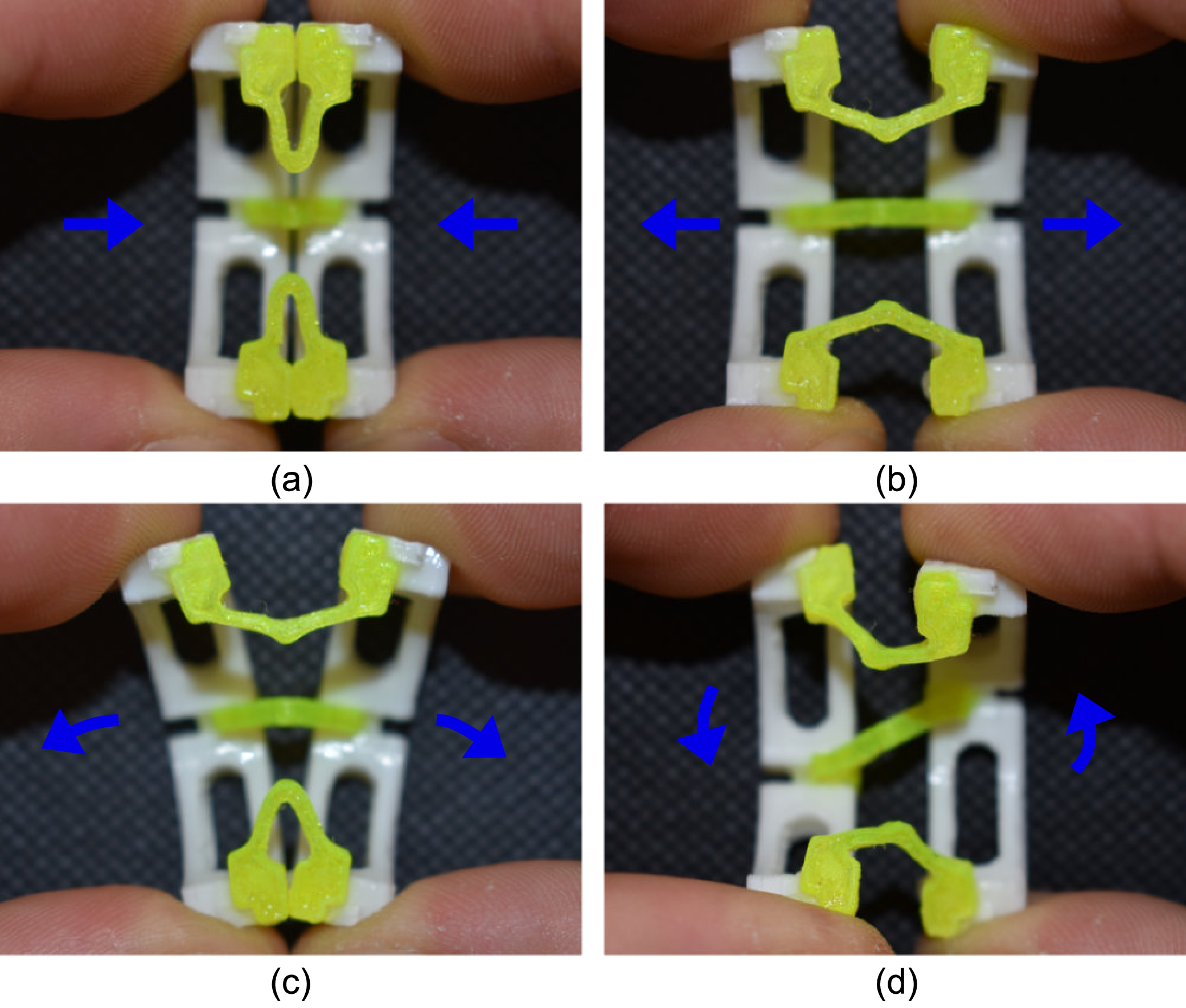}
    \caption{Motion of the flexible spine. Compression {\bf (a)}, extension {\bf (b)}, bending {\bf (c)}, and twist {\bf (d)}.}
    \label{SpineDOF}
\end{figure}

With the compression and expansion limits known, we can rearrange equation \eqref{central-gap} to solve for ligament beam length. The ligament neutral angle was set as 45\textdegree\ to allow for equal support to be generated at the stress concentrations of the ligament. The bending capability and 4 DOF of the spine is demonstrated in Fig.~\ref{SpineDOF}.

\subsection{Forming the Malleable Robot}
A robot manipulator with a very unique link component, malleable robots maybe require other unique components to function effectively. 

A 2-DOF malleable robot, formed from two revolute joints, a malleable link, and a rigid link, was developed following work in \cite{clark2020design}. Joint 1, positioned at the base of the robot, provides rotation in the z-axis. Joint 2 was positioned at the end of the malleable link, providing rotation in the axis perpendicular to the termination end. Both joints were constructed from a Dynamixel MX-64 servo motor, with a 3D printed ABS housing, and a thrust ball bearing (size 51106) providing force distribution of the motor torque to the output side of the joint. The malleable link is a variable stiffness link that uses Mylar-based layer jamming to transition between rigid and flexible modes, with a maximum length of 700~mm and a minimum length of 550~mm. The rigid link attached to the secondary joint has a length of 370 mm (actual distance of 450 mm between joint axis and end effector). The link was composed of a 42 mm~$\diameter$ polypropylene tube, and was attached to the robot using 3D printed ABS link ends similar to those used on the malleable link. The dimensions of the distal link were selected to be like that of the malleable link, with the shorter length chosen to prevent frequent collisions with the floor plane.

other
%

\subsection{Tutorial: Layer Jammed Malleable Link}
In this subsection we provide a tutorial for creating a single malleable link. In addition, commentary is provided at relevant steps for when considering modifications or replacements. The design utilises layer jamming as the variable stiffness technology, and a change in pneumatic pressure to activate (ambient to vacuum) and deactivate (vacuum to ambient) the stiffening.

\subsubsection{Preparation}
The necessary components and tools to follow this tutorial are detailed below, with the specific items used shown in brackets. Software versions used at the time of writing are noted. These specific items are only suggestions, and suitable alternatives or newer versions may/should be used instead.

\textbf{Material:}
\begin{itemize}
    \item 0.2mm Mylar \textsuperscript{\textregistered} (polyester film) sheets
    \item PETG filament - \textit{Prusament PETG}
    \item TPU filament - \textit{Sainsmart 95A TPU}
    \item 0.25mm sheet latex - \textit{T40 translucent smoky black 0.25mm}
    \item 4mm ID x 6mm OD PVC hose
    \item 0.6mm 135lb Braided fishing line
\end{itemize}
\textbf{Software:}
\begin{itemize}
    \item CAD package - \textit{SOLIDWORKS 2022-23}
    \item 3D printer slicing software - \textit{PrusaSlicer 2.5.0}
    \item Laser control software - \textit{Universal Laser System Control Panel}
\end{itemize}
\textbf{Tools:}
\begin{itemize}
    \item FDM 3D printer - \textit{Prusa MK3S}
    \item Laser cutter - \textit{Universal Laser System VLS 3.50}
    \item Cyanoacrylate glue - \textit{Gorilla Superglue Gel}
    \item Epoxy glue - \textit{Gorilla Epoxy}
    \item Copydex latex adhesive
\end{itemize}

\subsubsection{Method}
The instructions are broken down into sub-components, followed by a final assembly.

\textbf{Layer Jamming}
The layer jamming is the core component that enables the variable stiffness of the malleable link. It is achieved by overlapping multiple layers of material, here Mylar sheets, which are then sandwiched between two membranes. To provide a strong yet light structure, a tube is typically formed of which the length can be adjusted to suit the user’s needs. To form this tube, the flat sheets of Mylar must be cut in a conical design so that they wrap around the circumference of this tube.

\begin{figure}[t!]
    \centering
    \includegraphics[width=\columnwidth]{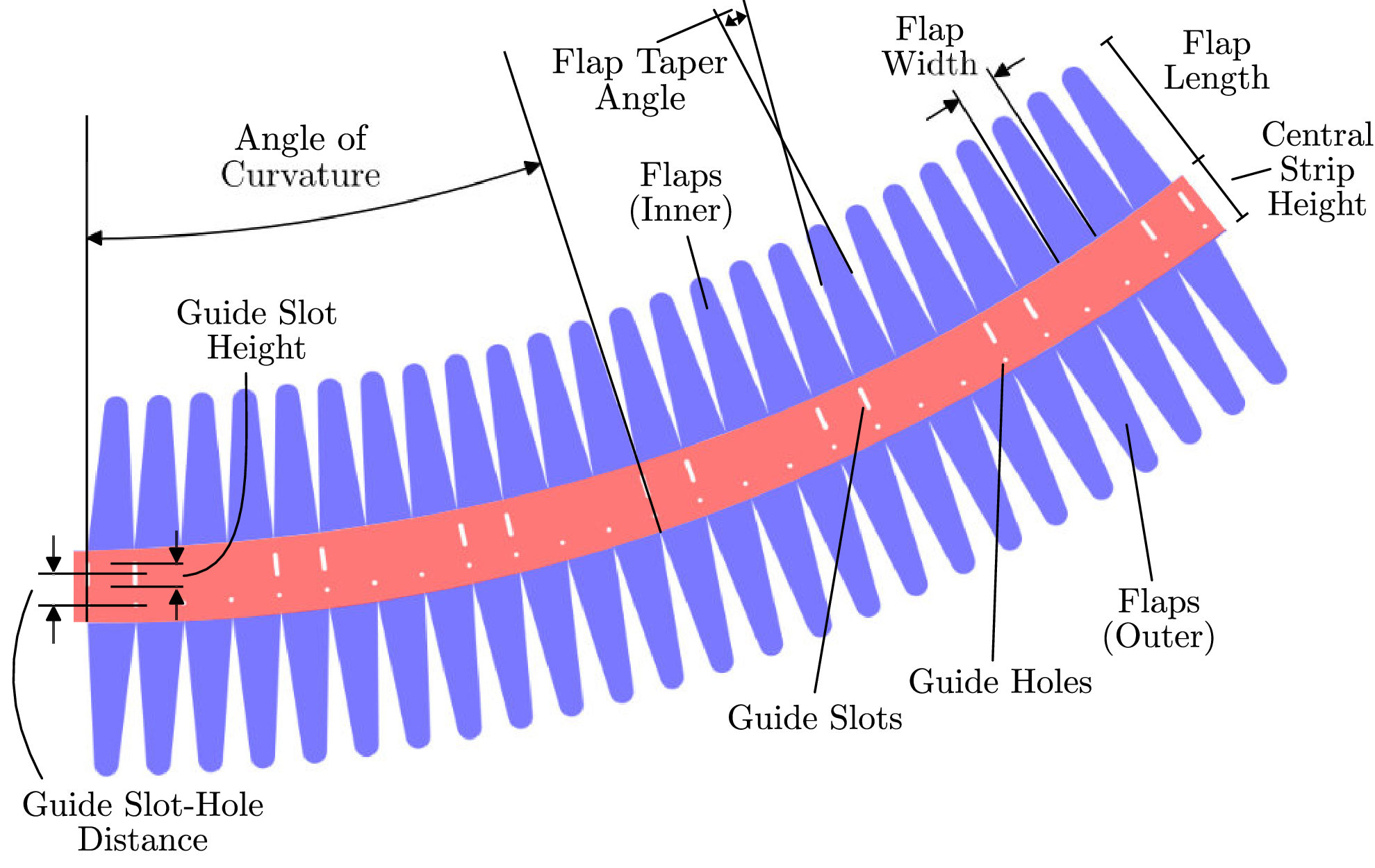}
    \caption{Diagram highlighting the key elements that compose the layer jamming structure.}
    \label{layer_structure}
\end{figure}

The basic structure of such a conical design is described in Fig.~\ref{layer_structure}. In consists of a central strip (shown in red) that is curved slightly upwards, and several flaps (shown in blue) that protrude outwards (vertically up and down) from the central strip. The angle of curvature of the central strip determines the diameter of the tube that is formed, whereas the length and width of the flaps determines the total frictional area, and therefore the overall maximum stiffness. One would think to therefore maximise this value, however an increase in the length and width of the flaps limits the bending ability due to restricted bending points, and it is partially limited by the taper of the flaps as above a certain length they are tapered to an insignificant width. Therefore, a healthy middle ground for the length of the flaps must be selected that balances the flexibility with the maximum stiffness. As this relates directly to the diameter of the tube that is formed and the personal preference in bending ability, it is not detailed here and must be selected by the user. That said, a value that produces 12 flaps per circumference has been shown to work well as a starting value.

The overlapping components of the design (the flaps) must be separated to ensure they do not overlap undesired parts of the design (they should only overlap other flaps directly above and below them, not to the side), and that they do not hinder the bending ability of the design. As the central strip is curved, this results in the flaps being tapered. Note, it is worth considering the manufacturing method here too, as the taper may need to be increased to ensure the flaps cleanly separate (e.g. in the case of laser cutting).

Ideally, a single, conical piece wrapped around would form the tube. This is typically not possible due to the significant size of the unwrapped conical shape, and therefore the design must be made modular. This is achieved using the guide slots and guide holes, which are used to align and join a following piece. In the design provided, two full circumferences is possible in a single piece (guide slots located every 120\textdegree ), after which another piece is required to join to continue wrapping. You should consider the tools available to you before the design process as the maximum size of the laser cutter will limit the upper bound of the layer jamming size you can create. 

\begin{figure}[t!]
    \centering
    \includegraphics[width=\columnwidth]{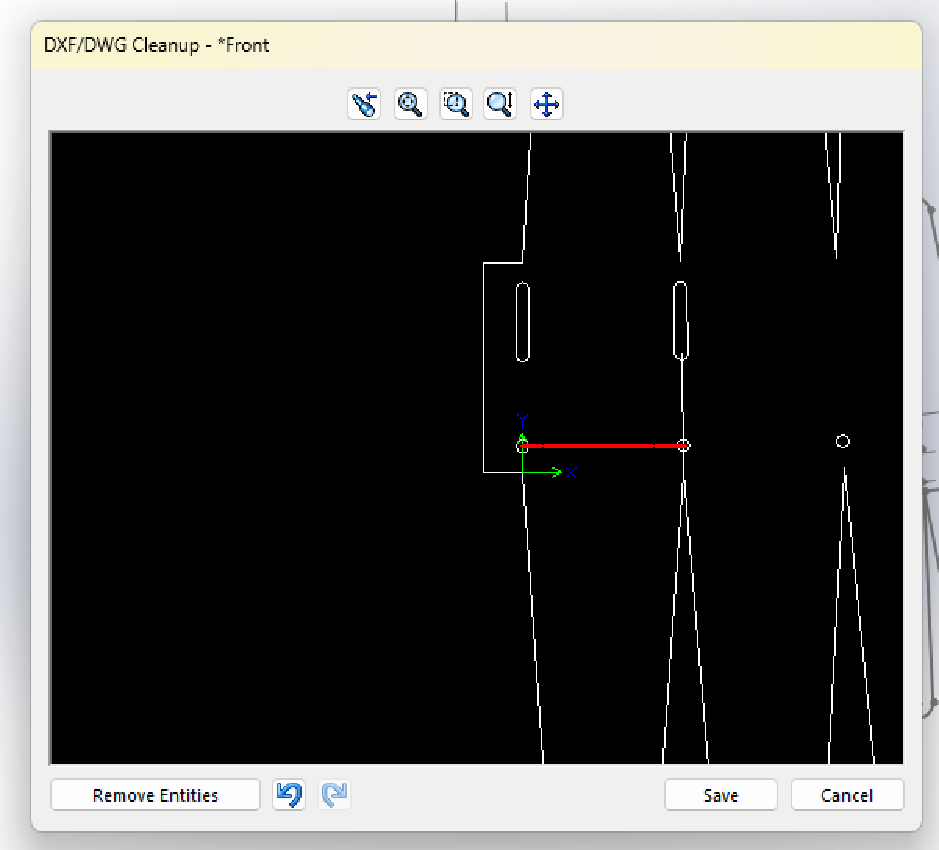}
    \caption{Cleanup options when exporting a DXF from SOLIDWORKS.}
    \label{DXF_Cleanup}
\end{figure}

To obtain an accurate design, it is suggested to use a 2D parametric design software to create the shape to cut out, here SOLIDWORKS is used (as a single sketch). This allows the use of equations and variables to define the shape, which once setup allow easy modification of the design. Once created, the design can then be exported as DXF, a 2D vectorised design. Note, make sure to only export the lines that you wish to cut, and ensure to remove items such as construction lines (in SOLIDWORKS you are provided with a menu to select the lines you would like to include when you export as DXF, see Fig.~\ref{DXF_Cleanup}). An example SOLIDWORKS part file is made available as a base design to start with and modify. The resulting exported DXF file is also made available (if any changes are made to the SOLIDWORKS file, a new exported DXF will need to be generated following the above instructions.).

According to the labelling in Fig.~\ref{layer_structure}, the provided CAD file and DXF have the initial dimensions shown in Table~\ref{layer_structure_values}, which can be modified easily using either the Equation Manager in SOLIDWORKS or directly.

\begin{table}[!t]
\caption{Specifications of Layer Jamming Pattern Parameters}
\label{layer_structure_values}
\begin{center}
\begin{tabularx}{.4\columnwidth}{lc}
\hline
Value               &   Dimension   \\
\hline
Angle of Curvature        &   11\textdegree    \\
Central Strip Height       &   16mm    \\
Flap Length       &   45mm    \\
Flap Taper Angle  &   8\textdegree     \\
Flap Width           &   12mm     \\
Guide Slot Height	&	5mm	\\
Guide Slot-Hole Distance	&	9.5mm	\\
\hline
\end{tabularx}
\end{center}
\end{table}

Once exported, it is now time for manufacture. The laser cutter is used for this task, cutting the Mylar sheet material. It is not possible to reliably cut multiple stacked sheets, therefore it is recommended to cut a single sheet at a time. Once the DXF file is imported to the laser software, ensure that the entire design is set to ‘CUT’, and not ‘ENGRAVE’ or ‘RASTER’. The laser cutter should come preloaded with presets for a variety of materials. For the ULS VLS 3.50 used here, there is a polypropylene/Mylar option which can be used to apply the correct speed/power settings of the laser to guarantee a clean cut. As the layers are modular, it is necessary to calculate the number of pieces needed to cut to achieve the desired tube length.

\begin{figure}[t!]
    \centering
    \includegraphics[width=\columnwidth]{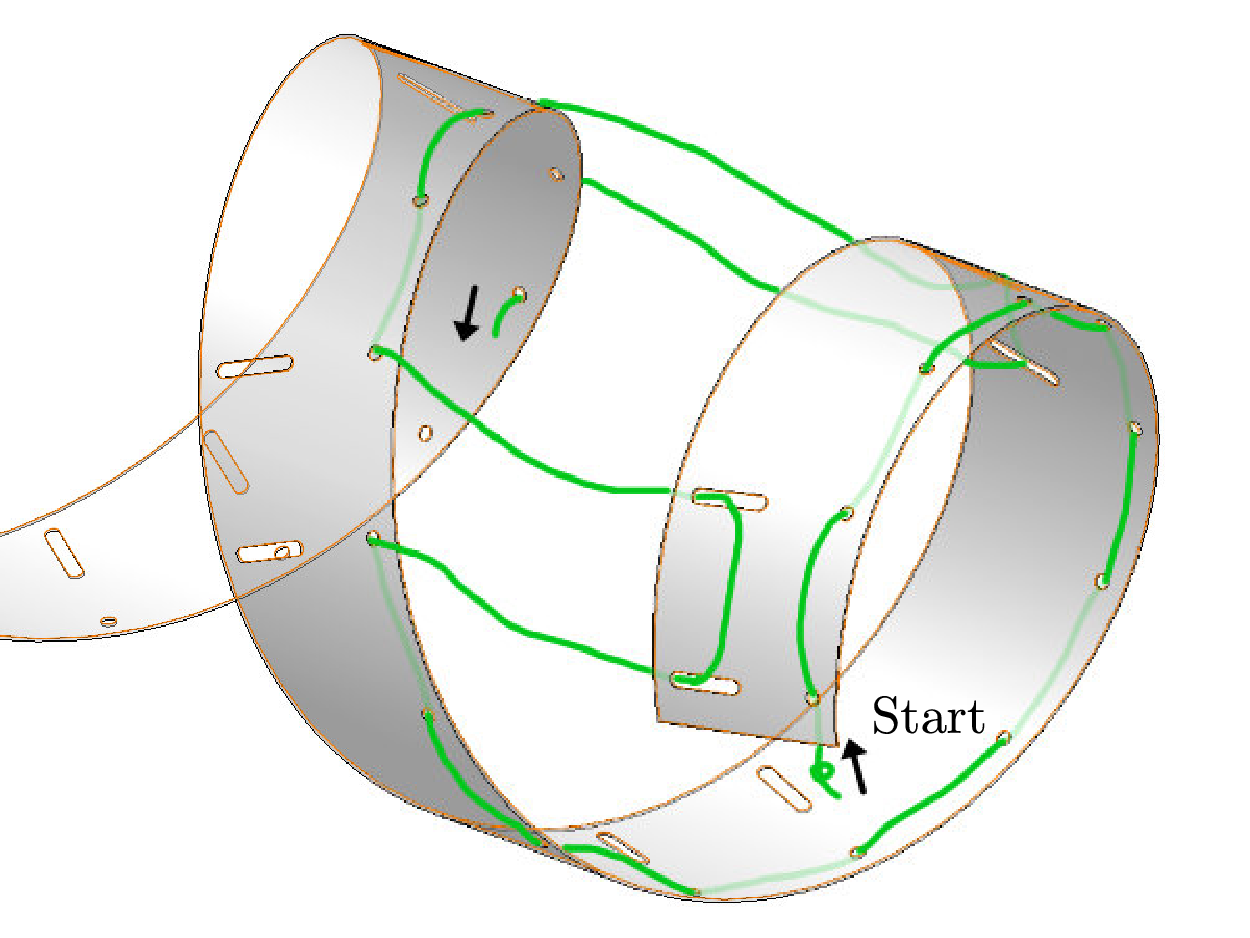}
    \caption{Diagram showing the sewing route (green) for the layer jamming. External flaps are not shown to increase visibility.}
    \label{sewing}
\end{figure}

Once cut, the final stage of the layer jamming is to assemble the layers from their 2D form into a 3D tube. This is achieved using either thread or a cable, which is sewn along the guide holes and guide slots around the circumference of the tube. Here, braided fishing line is used as it is extremely strong and quite easy to tie off. Wire has the advantage of increased strength, but limited flexibility and as such is more of a challenge during assembly. To form the 3D shape, the line is first fixed at one end of the bottom row of the guide holes. This is then sewn back and forth along the guide holes for a single circumference. The layers are then curved to the desired diameter, aligning the guide slots with the guide holes, and the sewing continues, ensuring that the line enters and exits at the inside of the layer tube for the guide slots. A diagram showing the line route is shown in Fig.~\ref{sewing}. Once complete, it should be tightened enough to secure the structure shape, while allowing the layers to move over each other. Finally, the additional flaps at the top and bottom of the cylinder can be removed (cut off) to provide a smooth flat surface at each end of the cylinder.

\textbf{Membranes}
To apply a vacuum to the layer tube, it must be sealed within a membrane. The easiest way to achieve this is by using two tubes, one larger and the other smaller than the layer tube. These can then be placed outside and inside the layer tube respectively, and when sealed at the ends form an airtight volume. The membranes must be flexible such that they do not hinder the bending curvature of the layer tube nor break under tension, and strong enough to resist the pressures of a high vacuum. From our testing, 0.25mm thick latex sheet is ideal. This can easily be formed into tubes of custom dimensions by joining two edges of a flat quadrilateral using latex glue (such as Copydex latex adhesive). When connecting the edges, it is imperative to ensure an airtight seal. Overlapping a large portion of the edges helps to achieve this.

\textbf{Supporting Spine}
With the layer tube as it is now, under extreme bending angles the structure will buckle and collapse in on itself. There is also a chance of buckling while in rigid mode when under extreme loads. To prevent this, a supporting spine structure is required down the centre of the tube. This supporting structure should have a rigid component (that resists the buckling) and a flexible component (that allows the support structure to follow the curvature of the layer tube). The layer tube can bend in 4 degrees of freedom: Bending in two dimensions, twisting, and extending/compressing. The supporting structure must also be capable of these.

\begin{figure}[t!]
    \centering
    \includegraphics[width=\columnwidth]{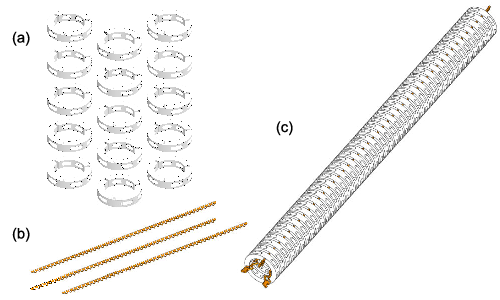}
    \caption{The parts of the spine: (a) the rigid component, (b) the flexible components, and (c) the assembled spine.}
    \label{spine}
\end{figure}

There are many designs that would satisfy the requirements, and one such design is presented here as an example. The design uses acrylonitrile butadiene styrene (ABS) as the rigid material, and thermoplastic polyurethane (TPU) as the flexible material. Both materials can be 3D printed, which eases manufacturing. Here, the rigid components are hollow discs, which provide a strong outer edge to support the layers, and a hollow centre to allow for items to pass through the length of the tube. This can be key for when integrating the malleable link into a larger design, such as a malleable robot that requires cables and tubes to run throughout it. The flexible TPU component is shaped with a zig-zag pattern such to prevent self-collisions when the tube bends, and to increase the ease of bending. The spacing of the rigid/flexible components is dependent on the layers bending radius. When printing using FDM printers, components are often weakest in the vertical Z axis. To ensure maximum strength of the TPU component, it is printed on its side. Once printed, the ABS and TPU components can be connected using an adhesive (such as superglue). The components along with the assembled spine are shown in Fig.~\ref{spine}. Using more advanced multi-material 3D printers, it is possible to avoid the assembly step by printing both components pre-assembled in a single print. Considerations must be taken here to ensure that the rigid and flexible components bond securely, and that the design still maintains the flexibility. For example, is the design printed in an expanded state (requiring a lot of support material), or in a compressed state? Using a dissolvable material for the support material could also provide advantages here, at an increased cost. The spine design shown here is made available to the reader in both editable CAD files (SOLIDWORKS, both part and assembly) and printable 3MF files. 

\textbf{Link Terminations}
With the components so far, we can make a tubular structure. However, this structure is currently open at both ends, not forming a sealed volume as needed. Also, the only exposed surface on the structure is a rubber membrane, which is difficult to form a strong, solid connection to. To correct these problems, we must introduce ‘malleable link terminations’. These are rigid components that are attached to each end of the tubular structure that provide a solid connection point for integrating the malleable link into a larger assembly and form an airtight seal on the structure. They also provide a singular location for all of the other components to connect together, as so far they are next to each other but not connected (such as the membrane and layers, or layers and spine). We must also have an access point to the internals of the sealed volume to evacuate it, which can be achieved using a pre-cut hole in the termination, into which a 6mm PVC tube can be inserted. Here, the terminations can simply be 3D printed out of the same rigid material used for the spine. For the design, it is key to have the outer membrane connect further down the termination than the layers/inner membrane, otherwise it is very difficult to form a complete seal. It is also key to consider the location of the tube/s to the sealed volume, as the radial distance between the inner and outer membrane and the membrane positioning will limit their size and placement. As with the other components, the design is made available to the reader in both editable CAD files (SOLIDWORKS) and printable 3MF files.

\textbf{Final Assembly}
The final step is to assemble everything together. Starting from the link termination with the vacuum tube insert, the PVC tube must be inserted and fixed in position (superglue). We can now assemble from the inside-out, starting with the spine. The spine is fixed to the termination using the pre-defined slots for the flexible components. These can also be fixed using the superglue. Ensure that only one of the two terminations is attached! The other must be left until later. Next, we slide the inner membrane over the spine, and fix it to the termination. Instead, this time we use the slower drying epoxy, which can be spread over the connection area on the termination before placing the membrane on it. Due to the translucency of the membrane, you can check the airtight seal is formed by ensuring that the dark band formed by the epoxy is seen in a continuous line around the entire circumference. Care must be always taken with both membranes to make sure no damage is caused to them, affecting the integrity of the seal. Once dry, we can now slide the layers over the inner membrane and fix them in place using more epoxy. Care must be taken here so that the epoxy is only applied to the desired area, and does not accidentally get into the rest of the layers, which once set would limit the flexibility of the link. Note that alternative methods to connect the layers could be taken here, such as using more thread or a clamping system. Once set, the outer membrane can be fixed in place using the same method as the inner membrane. Now all the components are fixed to one of the termination ends, we can attach the other. This must be fixed from the centre of the tube working outwards, using the necessary adhesives for each components.

You can check the assembly was successful by applying a vacuum down the expose 6mm tube, which should cause the link volume to evacuate and the link to transition to its rigid state. By removing the vacuum and returning the volume to ambient pressure, the link should transition to the flexible state. Here, the link should be manipulated in all four bending directions to confirm everything is fixed adequately, and nothing is restricting the motion of the link. If any problems arise, it is most likely an incomplete seal, which can often be fixed using more adhesive. If all the above tests are successful, congratulations! You have now assembled your own malleable link.


\section{Workspace computation and Kinematics computation}

\subsection{Overview of Distance Geometry}\label{sec:mr_dist}
Distance geometry, first defined by Leonard Blumenthal in first half of the 20th century \cite{blumenthal1953theory}, is a relatively new branch of mathematics that focuses on the study of geometries through the use of their metrics -- i.e. their distances. It avoids the need to define arbitrary reference frames, and allows for translations and rotations to be separately computed \cite{rojas-thesis}, compared to typical kinematic modelling using Denavit-Hartenburg (D-H) parameters. As such, distance geometry has successfully been used to intrinsically characterise Euclidean spaces \cite{havel1991use}. In this chapter, distance geometry is used as an alternative to other methods for computing the kinematics and topology reconfiguration, such as typical D-H parameters or Screw theory, as it avoids further hardware and software complexities. For example, it may be challenging to directly obtain the relative joint positions and angles using internal sensors in the malleable robot, which is an ongoing area of research in continuum robots \cite{wang2017soft}. In contrast, distance geometry only requires the localisation of points distributed along the robot, which can simply be achieved with external motion tracking. This is key for the computation of workspaces of malleable robots as traditional strategies based on the attachment of reference frames to the robot joints, such as those that make use of the D-H convention \cite{kung2005development, li2011design}, cannot be employed since both link dimensions and the relative orientation of the joints can change. An alternative is to perform the workspace analysis using screw theory \cite{coppola20146, xie2015design} or distance geometry \cite{rojas-thesis} as in these approaches the parameterisation does not depend on relative angles and distances between joint reference frames. We make use of a distance-geometry-based method herein as the technique has been shown to simplify the computation of the workspace equation of complex mechanisms \cite{rojas2016coupler, rojas-peaucellier}.

\subsubsection{Notation}
In what follows, we denote a point in $\mathbb{E}^3$ as $P_i$, $\mathbf{p}_{i,j}=\overrightarrow{\rm P_iP_j}$ denotes the vector from $P_i$ to $P_j$, $\mathbf{p}_{i,j,k}=\mathbf{p}_{i,j}\times\mathbf{p}_{i,k}$ denotes the cross product between vectors $\mathbf{p}_{i,j}$ and $\mathbf{p}_{i,k}$, and $s_{i,j}=\norm{\mathbf{p}_{i,j}}^2=d_{i,j}^2$ denotes the squared distance between $P_i$ and $P_j$, with vector coordinates arranged as column vectors. The vectors $\mathbf{p}_{i,j}$, $\mathbf{p}_{i,k}$, and $\mathbf{p}_{i,j,k}$ in general represent a non-orthogonal reference frame that is denoted by the column vector of nine components $\mathbf{q}_{i,j,k}=(\mathbf{p}_{i,j}^T, \mathbf{p}_{i,k}^T, \mathbf{p}_{i,j,k}^T)^T$.

\begin{figure}[t!]
    \centering
    \includegraphics[width=0.8\columnwidth]{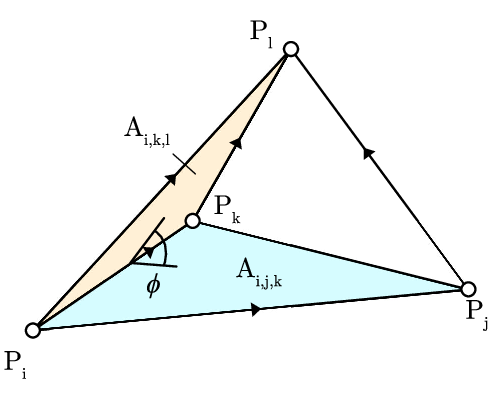}
    \caption{A dihedral angle $\phi$ of the tetrahedron $\pppp_{i,j,k,l}$ defined by the two triangles $\ppp_{i,j,k}$ and $\ppp_{i,k,l}$. Base vectors $\mathbf{p}_{i,j}$ and $\mathbf{p}_{i,k}$ and output vectors $\mathbf{p}_{i,l}$, $\mathbf{p}_{j,l}$, and $\mathbf{p}_{k,l}$ are shown. In this case, $V_{i,j,k,l}>0$.}
    \label{planeangle}
\end{figure}

The tetrahedron defined by points $P_i$, $P_j$, $P_k$, and $P_l$ is denoted as $\pppp_{i,j,k,l}$, with its \textit{origin} located at $P_i$, its \textit{base} given by the triangle $\ppp_{i,j,k}$ with area $A_{i,j,k}$, \textit{base vectors} $\mathbf{p}_{i,j}$ and $\mathbf{p}_{i,k}$, and \textit{output vectors} $\mathbf{p}_{i,l}$, $\mathbf{p}_{j,l}$, and $\mathbf{p}_{k,l}$. This notation is shown in Fig.~\ref{planeangle}~\cite{rojas2017distance, rojas2018forward}.

\subsubsection{Cayley-Menger Determinants}
The \textit{Cayley-Menger bideterminant} of two sequences of \textit{n} points, $[P_{i,1},...,P_{i,n}]$ and $[P_{j,1},...,P_{j,n}]$, is defined as \cite{rojas-thesis}

\small
\begin{equation*}
    D(i_1,...,i_n;j_1,...,j_n)=2\left(\frac{-1}{2}\right)^n\left| \begin{array}{cccc}
                0       &   1               & \cdots    & 1             \\
                1       &   s_{i_1,j_1}     & \cdots    & s_{i_1,j_n}   \\
                \vdots  &   \vdots          & \ddots    & \vdots        \\
                1       &   s_{i_n,j_1}     & \cdots    & s_{i_n,j_n}   \\
                \end{array} \right|.
\end{equation*}
\normalsize

When the two point sequences are the same, $D(i_1,...,i_n;i_1,...,i_n)$, this is abbreviated as $D(i_1,...,i_n)$, known as the \textit{Cayley-Menger determinant}. For example, for the 5 points $D(P_1,\ldots,P_5)$ this is

\small
\begin{equation}
    \begin{split}
    D(1,2,3,4,5)&=-\frac{1}{16}\left| \begin{array}{cccccc}
                0   &   1         & 1       & 1       & 1       & 1       \\
                1   &   0         & s_{1,2} & s_{1,3} & s_{1,4} & s_{1,5} \\
                1   &   s_{1,2}   & 0       & s_{2,3} & s_{2,4} & s_{2,5} \\
                1   &   s_{1,3}   & s_{2,3} & 0       & s_{3,4} & s_{3,5} \\
                1   &   s_{1,4}   & s_{2,4} & s_{3,4} & 0       & s_{4,5} \\
                1   &   s_{1,5}   & s_{2,5} & s_{3,5} & s_{4,5} & 0       \\
                \end{array} \right|.
    \end{split}
    \label{eq-CMD}
\end{equation}
\normalsize

For the general point sequence $P_{1}$, $P_{2}$,$\ldots$,$P_{n}$, the Cayley-Menger determinant gives $(n-1)!^2$ times the squared hypervolume of the simplex spanned by the points in $\mathbb{E}^{n-1}$ \cite{Menger}. Hence, $D(1,2,3,4,5) = 0$ in $\mathbb{E}^3$. Similarly, for $n=3$, we have \cite{thomas2005revisiting}, 

\begin{equation}
    D(i,j,k)=4A_{i,j,k}^2=\norm{(P_j-P_i)\times(P_k-P_i)}^2,
    \label{eq1}
\end{equation}

which is the Heron's formula relating the area $A_{i,j,k}$ of triangle $\ppp_{i,j,k}$. This can also be expressed purely in interpoint distances as:
\begin{multline}
    A_{i,j,k}=(p_{i,j,k}(p_{i,j,k}-\norm{\textbf{p}_{i,j}})\\(p_{i,j,k}-\norm{\textbf{p}_{i,k}})(p_{i,j,k}-\norm{\textbf{p}_{j,k}}))^\frac{1}{2},
    \label{eq-tri-area}
\end{multline}
where $p_{i,j,k}$ is half the perimeter of the triangle $\ppp_{i,j,k}$ defined as
\begin{equation}
    p_{i,j,k}=\frac{1}{2}\left(\norm{\textbf{p}_{i,j}}+\norm{\textbf{p}_{i,k}}+\norm{\textbf{p}_{j,k}}\right).
\end{equation}

For $n=4$ we obtain the orientated volume $V_{i,j,k,l}$ of the tetrahedron $\pppp_{i,j,k,l}$ as

\begin{equation}
    D(i,j,k,l)=36V_{i,j,k,l}^2.
    \label{eq2}
\end{equation}
It is defined as positive if $\abs{\mathbf{p}_{i,j},\mathbf{p}_{i,k},\mathbf{p}_{i,l}}>0$, and negative otherwise \cite{rojas2017distance}. For \textit{Cayley-Menger bideterminants}, for $n=3$ we have 

\begin{align}
    D(i,j,k;i,k,l)&=4A_{i,j,k}A_{i,k,l}\cdot\cos{(\phi_{i,j,k,l})} \nonumber \\
    &=((P_i-P_k)\times(P_j-P_k))\cdot \nonumber\\
    &\,\,\,\,\,\,\,\,((P_i-P_l)\times(P_k-P_l)),
    \label{eq3}
\end{align}
where $\phi_{i,j,k,l}$ is the dihedral angle between the two planes defined by the triangles $\ppp_{i,j,k}$ and $\ppp_{i,k,l}$. This can be seen in Fig.~\ref{planeangle}.

\subsubsection{Trilateration}
Trilateration is a method for computing the location of an unknown point using known distances of the point from 3 different known sites. For example, given a tetrahedron $\pppp_{i,j,k,l}$ (Fig.~\ref{planeangle}), we can compute the output vector $\textbf{p}_{i,l}$ as \cite{rojas2017distance}

\begin{equation}
    \textbf{p}_{i,l} = \textbf{W}_{i,j,k,l}\textbf{q}_{i,j,k},
    \label{eq-w}
\end{equation}
where

\begin{equation}
    \textbf{W}_{i,j,k,l}^T=\frac{1}{4A_{i,j,k}^2}\left(\begin{array}{c}
    -D(i,j,k;i,k,l)\textbf{I}\\
    D(i,j,k;i,j,l)\textbf{I}\\
    6V_{i,j,k,l}\textbf{I}
    \end{array}\right),
    \label{eq-w2}
\end{equation}
with \textbf{I} being the $3\times3$ identity matrix.

\subsection{Workspace Definition}\label{sec:mr_work}
The workspace of a robot arm is defined as the region (or surface/volume) within which every point can be reached by the end effector, and is one of the most important specifications for both robot designers and users \cite{tsai-algorithm}. We can either compute the workspace given the structure (analysis), or alternatively determine the robot structure from a desired workspace (synthesis) \cite{gupta-design}. For the case of malleable robots, traditional strategies for the computation of workspaces based on the attachment of reference frames to the robot joints, such as those that make use of the Denavit-Hartenberg convention \cite{kung2005development, li2011design}, cannot be employed since both link dimensions and the relative orientation of the joints can change. An alternative is to perform the workspace analysis using screw theory \cite{coppola20146, xie2015design} or distance geometry \cite{rojas-thesis} as in these approaches the parameterisation does not depend on relative angles and distances between joint reference frames. We make use of a distance-geometry-based method herein as the technique has been shown to simplify the computation of the workspace equation of complex mechanisms \cite{rojas2016coupler, rojas-peaucellier}.

\begin{figure}[t!]
    \centering
    \includegraphics[width=\columnwidth]{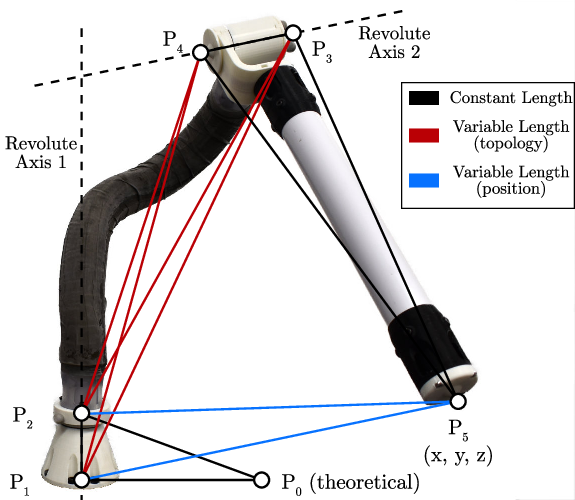}
    \caption{The 2-DOF malleable robot arm can be modelled as a bar-and-joint framework formed by connecting 5 points: $P_1$ and $P_2$, which define the first axis; $P_3$ and $P_4$, which define the second axis; and $P_5$, which corresponds to the centre of the end effector. An additional offset point $P_0$ is added, which is used to define the angle of the first axis. The distances between points can then be categorised as constant distances that do not change (\textbf{black}), distances that vary based on the topology configuration of the robot (\textbf{red}), and distances that vary based on the joint positioning of the robot (\textbf{blue}).}
    \label{distancegeometry}
\end{figure}

We define our 2-DOF malleable robot with a vertical rotary joint at the base, connected co-linearly to one end of a malleable link, which terminates at a second rotary joint mounted perpendicularly to the other end of the malleable link. A second, rigid link is then attached to the second rotary joint, also perpendicularly, which then terminates at the end effector. Since a link connecting two skew revolute axes can be modelled as a tetrahedron by taking two points on each of these axes and connecting them all with edges, and a rigid link connected to a revolute axis can be modelled as a triangle by taking two points on the axis and a point at the end of the link and connecting them all with edges \cite{rojas2016coupler}, we can model the developed 2-DOF malleable robot using distance geometry as a bar-and-joint framework of 6 points and 12 edges, shown in Fig.~\ref{distancegeometry}, with $P_5$ corresponding to the end-effector, and $P_1$ corresponding to the robot origin. The axes of the revolute joints are defined by the points $P_1$ and $P_2$ for joint 1, and $P_3$ and $P_4$ for joint 2. An additional fixed point $P_0$ is defined offset from the origin at $P_1$, which is necessary for the forward and inverse kinematics, but not for the workspace definition or topology reconfiguration. The interpoint distances can further be categorised as:
\begin{enumerate}
  \item Distances with constant length, that do not change with variation in robot topology and positioning ($\mathbf{p}_{0,1}$, $\mathbf{p}_{0,2}$, $\mathbf{p}_{1,2}$, $\mathbf{p}_{3,4}$, $\mathbf{p}_{3,5}$, $\mathbf{p}_{4,5}$).
  \item Distances that vary with changes in robot topology, but not robot positioning ($\mathbf{p}_{1,3}$, $\mathbf{p}_{1,4}$, $\mathbf{p}_{2,3}$, $\mathbf{p}_{2,4}$).
  \item Distances that vary with changes in robot positioning, but not robot topology ($\mathbf{p}_{1,5}$, $\mathbf{p}_{2,5}$).
\end{enumerate}

For defining the workspace, we can represent this as a Cayley-Menger determinant of 5 points (not including $P_0$), shown in Eq.~\ref{eq-CMD}. Since $D(1,2,3,4,5) = D(4,3,2,1,5) = 0$, using properties of the determinant of block matrices \cite{Powell}, it can be shown that this condition can be compactly expressed using $3\times3$ matrices as
\begin{align}
    D(1,2,3,4,5) = 2\,s_{1,2}\,s_{1,5}\,s_{2,5}\det(\mathbf{A}-\mathbf{B}\mathbf{C}\mathbf{B}^T)=0, \label{eq:determinant2}
\end{align}
where
\begin{align*}
    \mathbf{A} &= \left[ \begin{array}{ccc}
                0   &   1         & 1       \\
                1   &   0         & s_{3,4} \\
                1   &   s_{3,4}   & 0       \\
                \end{array} \right],\, 
    \mathbf{B} = \left[ \begin{array}{ccc}
                1   &   1         & 1       \\
                s_{2,4}   &   s_{1,4}         & s_{4,5} \\
                 s_{2,3}   &   s_{1,3}         & s_{3,5} \\
                \end{array} \right], \textrm{ and} \\           
    \mathbf{C} &= \frac{1}{2}\left[ \begin{array}{ccc}
                -\frac{s_{1,5}}{s_{1,2}\,s_{2,5}}   &  \frac{1}{s_{1,2}}         &    \frac{1}{s_{2,5}}    \\
                \frac{1}{s_{1,2}}   &   -\frac{s_{2,5}}{s_{1,2}\,s_{1,5}}         & \frac{1}{s_{1,5}} \\
                 \frac{1}{s_{2,5}}   &   \frac{1}{s_{1,5}}         & -\frac{s_{1,2}}{s_{1,5}\,s_{2,5}} \\
                \end{array} \right].
\end{align*}

\begin{figure*}[t!]
    \centering
    \includegraphics[width=\textwidth]{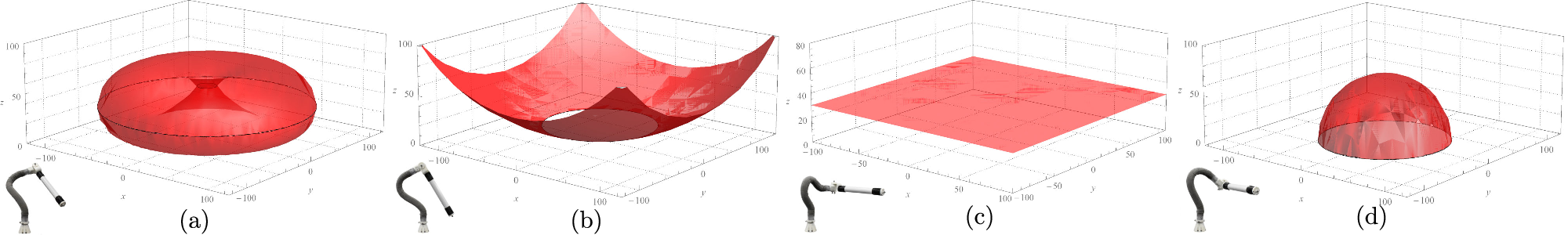}
    \caption{Simulated example workspaces for each type of robot topology achievable by a 2-DOF malleable robot: \textbf{(a)} Spherical, \textbf{(b)} PUMA-like, \textbf{(c)} SCARA, and \textbf{(d)} General Articulated. \cite{clark2020design}}
    \label{simulationworkspace}
\end{figure*}

Following the notation of Fig.~\ref{distancegeometry}, equation \eqref{eq:determinant2} is solely satisfied in the points in $\mathbb{E}^{3}$ where a 2-DOF malleable robot can physically exist. This fact can be exploited to compute the Cartesian equation of the robot workspace, say $\Gamma(x, y, z)$, by deriving the locus of point $P_5$, the end effector, the coordinates of which are $\mathbf{p}_5 = (x, y, z)$ in a particular reference frame. To simplify this computation we can assume, without loss of generality, that $P_1$ equals the origin of the global reference frame and that $P_2$ is located in the positive side of the $z$-axis, such that $\mathbf{p}_1 = (0, 0, 0)$ and $\mathbf{p}_2 = (0, 0, d_{1,2})$. Therefore,
\begin{align}
\begin{split}
    s_{1,5} &= x^2 + y^2 + z^2\\
    s_{2,5} &= x^2 + y^2 + z^2 - 2d_{1,2}z + s_{1,2}. \label{eq:basedistances}
\end{split}
\end{align}

Substituting equation \eqref{eq:basedistances} into equation \eqref{eq:determinant2}, fully expanding the result and rearranging terms, we get
\begin{align}
    \Gamma(x, y, z) & \myeq  q_0\,(x^2+y^2+z^2)^2+q_1\,d_{1,2}\,z\,(x^2+y^2+z^2) \nonumber \\
    & q_2\,x^2+q_2\,y^2+q_3\,z^2+q_4\,d_{1,2}\,z+q_5, \label{eq:Gamma}
\end{align}
where $q_i,\,i=0,\ldots,5$ are polynomials in $s_{1,2}=d_{1,2}^2$, $s_{1,3}$, $s_{1,4}$, $s_{2,3}$, $s_{2,4}$, $s_{3,4}$, $s_{3,5}$, and $s_{4,5}$. $\Gamma(x, y, z)$ is an algebraic surface of degree 4 (a quartic surface) that corresponds to the workspace surface, traced by the end effector (point $P_5$), of a 2-DOF malleable robot. The expressions of the polynomials $q_i$ are not included here due to their length, however these polynomials can be easily reproduced using a computer algebra system following the steps given above.

By providing constraints to the two revolute axes of the malleable robot, we can define certain workspace categories belonging to specific robot configurations (topologies). Malleable robots are a general purpose serial robot, and so follow similar applications where the task workspace defines the configuration. The robot configurations we define are spherical, PUMA-like, SCARA, and general articulated. The constraints for each of them are discussed next.

\subsubsection{Spherical (or variable radius) case}
In a spherical robot configuration, the two revolute axes of the robot coincide at the base, such that, according to the notation of Fig.~\ref{distancegeometry}, points $P_1$ and $P_3$ are coincident. Thus, $s_{1,3} = 0$, $s_{2,3} = s_{1,2}$, and $s_{3,4} = s_{1,4}$. Substituting these values into \eqref{eq:Gamma}, we obtain
\begin{align}
    \Gamma_A(x, y, z) \myeq x^2 + y^2 + z^2 - s_{3,5}=0,
\end{align}
which corresponds to the equation of a sphere of radius $d_{3,5}$ centred at $P_1$. Observe that in this case the radius $d_{3,5}$ is not constant, it can be adjusted according to need. An example of this workspace can be seen in Fig.~\ref{simulationworkspace}(a).

\subsubsection{PUMA-like (or variable centre and radius) case}
In a PUMA-like robot configuration, the two revolute axes of the robot are perpendicular and coincide at a point located in the positive side of the $z$-axis, such that points $P_2$ and $P_4$ are coincident, and the angle $\angle P_1P_2P_3$ is $\frac{\pi}{2}$. Thus, $s_{2,4} = 0$, $s_{1,4} = s_{1,2}$, $s_{3,4} = s_{2,3}$, and $s_{1,3} = s_{1,2}+s_{2,3}$. Substituting these values into \eqref{eq:Gamma}, we get
\begin{align}
    \Gamma_B(x, y, z) \myeq x^2 + y^2 + (z-d_{1,2})^2-s_{4,5}=0,
\end{align}
which corresponds to the equation of a sphere of radius $d_{4,5}$ centred at $P_2$. Observe that in this case the centre $(0, 0, d_{1,2})$ and radius $d_{4,5}$ are not constant, they can be adjusted according to need. The same equation is obtained when the perpendicularity of the two axes is relaxed. An example of such a workspace can be seen in Fig.~\ref{simulationworkspace}(b).

\subsubsection{SCARA (or planar) case}
In a SCARA robot configuration, the two revolute axes of the robot are parallel. Using projective geometry arguments, this implies that there exist a point in the second axis, say $P_3$, such that the distance between it and the $xy$-plane is $\delta$, with $\delta>0$, $\delta \to \infty$. Hence, $d_{1,3}=\delta$, $d_{2,3}=d_{1,2}+\delta$, $d_{3,4}=z_{4}+\delta$, $d_{3,5}=z_{5}+\delta$, being $z_{i}$ the distance between $P_i$ and the $xy$-plane. Substituting these values into \eqref{eq:Gamma}, we obtain an equation that can be written as a quadratic polynomial in $\delta$, say $\Omega = k_2(x,y,z)\delta^2+k_1(x,y,z)\delta+k_0(x,y,z) = 0$. By factoring out $\delta^2$ in this polynomial, we get $\Omega =\delta^2(k_2(x,y,z)+\frac{k_1(x,y,z)}{\delta}+\frac{k_0(x,y,z)}{\delta})=0$. Since $\delta \to \infty$, then $\Omega = k_2(x,y,z)=0$. 

Since the two revolute axes of the  robot are  parallel, we have to include additional constraints in $\Omega=k_2(x,y,z)=0$, that is, $P_2$=$P_4$=$P^\infty$. This implies that $s_{2, 4} = 0$ and $d_{1, 4} = d_{1, 2}$. Substituting these values into $\Omega=k_2(x,y,z)=0$, we get $(z_4-d_{1, 2})\,s_{1, 2}\,\Phi(x,y,z) = \Phi(x,y,z) = 0$. We can then include the final constraint $z_4 = d_{1, 2}$ (as $P_2$=$P_4$) in the result ($\Phi(x,y,z)$). This yields,
\begin{equation*}
     \left( z-{\it z_5} \right)  \left( {x}^{2}+{y}^{2}+{z}^{2}-2\,d_{{1,2}
}z+{d_{{1,2}}}^{2}-s_{{4,5}} \right) = 0.
\end{equation*}
Following a similar procedure in the above equation to that done for $\delta$, but in this case for $d_{1,2}$ ($d_{1,2} \to \infty$ since $P_2=P_4=P^\infty$), we finally get
\begin{align}
    \Gamma_C(x, y, z) \myeq \left( z-{\it z_5} \right)=0, \label{eq:GammaSCARA}
\end{align}
which corresponds to the equation of a plane parallel to the $xy$-plane. Observe that $z_5$, the distance between the end effector and the $xy$-plane, is not constant and can be adjusted according to need. An example of this workspace can be seen in Fig.~\ref{simulationworkspace}(c).

\subsection{General articulated}
We define the general articulated robot configuration as any robot configuration that does not comply with any of the constraints of the 3 other defined robot configurations, thus the form of the workspace surface in this case is $\Gamma(x, y, z)=0$ (equation \eqref{eq:Gamma}). An example of this workspace, which corresponds to a torus, can be seen in Fig.~\ref{simulationworkspace}(d).

\subsection{Forward Kinematics}
In computing the forward kinematics of the malleable robot, we specify the joint angles (in this case, the dihedral angles--use of the physical joint values is addressed later), and obtain the new position of $P_5$.

We assume the robot is in a fixed topology, and that the current point positions (and therefore their interpoint distances) are known. Starting from the origin, we define $\phi_{1,0,2,3}$ as the joint 1 dihedral angle between the fixed triangle $\ppp_{0,1,2}$ and the current topology defined triangle $\ppp_{1,2,3}$. We then define $\phi_{3,2,4,5}$ as the joint 2 dihedral angle between the current topology defined triangle $\ppp_{2,3,4}$ and the constant length triangle $\ppp_{3,4,5}$.

We first compute the new location of P3, defined by the new dihedral angle $\phi_{1,0,2,3}$ value. Using the interpoint distances we can calculate the areas of both triangles ($A_{0,1,2}$,$A_{1,2,3}$) using Eq.~\ref{eq1}. Substituting into Eq.~\ref{eq3}, we can solve for the new distance $\mathbf{p}_{0,3}$. With all distances for the tetrahedron $\pppp_{1,0,2,3}$ known, we can compute the new position of $P_3$ using trilateration.

After using Eq.~\ref{eq2} to compute the orientated volume $V_{1,0,2,3}$, we can use the now known tetrahedron $\pppp_{1,0,2,3}$ distances to compute the new position of point $P_3$:

\begin{equation}
    P_3 = P_1+\textbf{W}_{1,0,2,3}\textbf{q}_{1,0,2}.
    \label{eq-p3}
\end{equation}

With the new position of $P_3$ known, and the constant distance $\textbf{p}_{3,4}$, we can compute the position of $P_4$ using the points $P_1$, $P_2$, and $P_3$ and their known interpoint distances:

\begin{equation}
    P_4 = P_1+\textbf{W}_{1,3,2,4}\textbf{q}_{1,3,2}.
    \label{eq-p4}
\end{equation}

Finally, we can compute the new location of $P_5$ using the same process for $P_3$, using the new positions of $P_3$ and $P_4$ and the new dihedral angle $\phi_{3,2,4,5}$ for joint 2:

\begin{equation}
    P_5 = P_3+\textbf{W}_{3,2,4,5}\textbf{q}_{3,2,4}.
    \label{eq-p5}
\end{equation}

As the orientation of the volume of tetrahedra requires all distances to be known (Eq.~\ref{eq2}), in determining $P_3$ and $P_5$ we instead define the orientation based on the dihedral angles, where it is defined as positive if $\phi_{i,j,k,l}<180^{\circ}$ and negative otherwise, where $0^{\circ}\leq\phi_{i,j,k,l}\leq360^{\circ}$.

\subsection{Inverse Kinematics}
For the inverse kinematics of the robot, we provide the end effector position ($P_5$) and compute the required dihedral angles necessary to obtain it. It is assumed the current topology of the robot (its positions and interpoint distances) are all known.

We first calculate the angle of joint 2, $\phi_{3,2,4,5}$. Using the new position of $P_5$, we know the new distance $\textbf{p}_{2,5}$. Rearranging Eq.~\ref{eq3} to solve for $\phi_{i,j,k,l}$ we obtain

\begin{align}
    \cos{(\phi_{i,j,k,l})}&=\frac{D(i,j,k;i,k,l)}{4A_{i,j,k}A_{i,k,l}},
    \nonumber \\
    &=\frac{D(i,j,k;i,k,l)}{D^\frac{1}{2}(i,j,k)D^\frac{1}{2}(i,k,l)}.
    \label{eq-phi2}
\end{align}

Using Eq.~\ref{eq-phi2} along with Eq.~\ref{eq-tri-area}, we can calculate the angle of joint 2 $\phi_{3,2,4,5}$ using only interpoint distances. This procedure returns the smallest value of the dihedral angle, limited to the range $0^{\circ}\leq\phi_{i,j,k,l}\leq180^{\circ}$. We can extend this value to the full $360^{\circ}$ range to find the two valid values of the dihedral angle, where it is $360^{\circ}-\phi_{i,j,k,l}$ or $\phi_{i,j,k,l}$ otherwise. These are known as the \textit{elbow up} and \textit{elbow down} configurations of an arm.

Next, we can compute the new position of $P_3$ as performed in the forward kinematics using Eq.~\ref{eq-w}, using the current topology points of the robot, along with the new position of $P_5$. Thus,

\begin{equation}
    P_3 = P_5+\textbf{W}_{5,1,2,3}\textbf{q}_{5,1,2}.
    \label{eq-ik-p3}
\end{equation}
With the new position of $P_3$ known, we can repeat the procedure for calculating $\phi_{3,2,4,5}$ for $\phi_{1,0,2,3}$. If desired, the new location of $P_4$ can also be computed using Eq.~\ref{eq-w} and the calculated dihedral angle $\phi_{1,0,2,3}$.


\section{Augmented reality for collaborative reconfiguration}

\subsection{Introduction} \label{Introduction}

The malleable robot, as introduced in the previous sections, offers a versatile and robust solution for a new class of low-cost reconfigurable robots. With the ideation of the malleable link, enabled through layer jamming, the user is offered the option of re-configuring the link into wide range of operation shapes, hence eliminating the need for designing a new robotic system in order to satisfy the new task. This results in a low DOF robot having a significantly increased task versatility, due to the variable workspace of the robot. Naturally, this raises a new set of challenges. For extrinsic malleable robots, where their reconfiguration is performed externally by reshaping the malleable link, the user must understand how to do so such that the resulting position allows the robot to perform a task. This involves conducting path planning and motion control without the help of classical forward and inverse kinematics frameworks, while also accounting for the presence of a human operator, and work towards an automated positioning system. 

Human-in-the-loop robotic assistance and control, or otherwise commonly known as a form of human robot interaction, is a well-studied field that acts as an important cornerstone for semi-autonomous robotic systems. Having originated in the 1980s with the emergence of intelligent robots and artificial intelligence \cite{dautenhahn2007socially}, HRI systems were explored for industrial automation and mobile robotics under different settings, paving the way for modern day systems. 30 years later, researchers began to incorporate human like features into such systems, including facial expressions in the case of humanoid and assistive robots, as well as gestures and spatial sound in the case of visual assisted systems \cite{makhataeva2020augmented}. In recent years, HRI research has continued to evolve, with a focus on developing robots that can adapt to the needs and preferences of individual users. Additionally, there has been increasing interest in the development of collaborative robots (cobots) that can work alongside humans in various settings. However, despite the progress made, under an industrial setting, robots still excel at performing precise, accurate and repetitive tasks while being kept away from humans, since it is difficult to visualise their intentions in real time and communicate that to a nearby user.
In this section, we will first explore this challenge from the perspective of HRI and investigate the available tools from literature that are suitable to us, as well as any necessary innovation aspects that are needed, then we will provide a guided walkthrough of the development process.

\begin{figure}[t!]
    \centering
    \includegraphics[width=80mm]{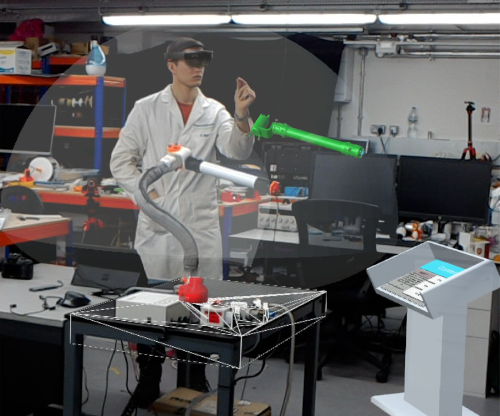}
    \caption{The developed augmented reality-assisted reconfiguration system of a malleable robot. The user is shown interacting with the AR assisted points placement scene overlaid on top of the robot base, with the presence of a transparent configuration space. The lectern to the right prints out the instructions for the task the user needs to conduct.}
    \label{main}
\end{figure}

\subsection{Human Robot Interaction}

In this work, we have focused on the development of a physical human-robot collaboration system, which is defined as the situation where humans, robots and the environment come into contact in order to accomplish a task, forming a tightly coupled dynamic system. What is important about this definition is that each component must be able to observe, and estimate the contribution of the other counter parts through sensory information. Once the changes in the environment is understood, the system then reacts to the change or enhances the observation to improve the collective performance \cite{ajoudani2018progress}. 
One of the most common HRI systems take the form of fusing assistive environment or cues with the reality, and suitable reactions from the system can be provided. Applications of the fusion system is especially important in safety critical applications, such as during maintenance or surgery.  In Neges et al.’s work \cite{neges2017combining}, the team utilised natural markers observed from the maintenance scene in order to better assist engineers and perform indoor navigation, thus significantly improving the efficiency of the work performed. Other works by Lee et al. \cite{lee2008product} has integrated augmented reality (AR) with facility management system, thus providing real time monitoring information of critical components in a building. Other examples include robotic surgery, where collaborative systems have been developed to assist with registration between the surgical scene and the medical instruments. For example, in the realm of orthopaedic surgery, Iqbal et al.  \cite{iqbal2021augmented} demonstrated a precise AR registration system that provides holographic feedback to the user depending on their position of the tool, where a guidance cylinder was used to provide the user with feedback on the quality of their tool alignment.  
On the other hand, HRI system may be control-based, where a specialised controller is designed to help the user in a passive or active manner in order to collaboratively achieve the goal. Notably, active control may take the form of feedforward \cite{magnanimo2014bayesian}, where the system predicts the user’s intentions, or feedback, which reacts to inputs of the system and minimises the error between the expectation and reality. An example of HRI in the form of dynamic active constraints can be found in Bowyer et al.’s work, where passive control was used to monitor the user’s actions, while providing a feedback to the user when a condition is met (eg. The user is deviating from a predefined metric \cite{bowyer2013active}, such as virtual boundaries). Despite its widespread adoption in the field, such algorithms were not adopted in our designed system, thus will not be discussed further. However, the reader is encouraged refer to \cite{bowyer2013active} for more details.

As the malleable robot does not possess any intrinsic actuation, the study of HRI will primarily focus on evaluating solutions for human supervisory control \cite{sheridan2016human}. With that said, most researchers agree that there needs to be a shared understanding of the robot's workspace in order to initiate the configuration process, such that effective collaboration can take place. Although this method of interaction is still novel, a study by Solyman et al. \cite{solyman2020perceptive} with 2D workspace visualisation has demonstrated the potential of these workflows to accelerate robot assembly tasks. The system provides an interface that allows the user to define objects in the scene to move, perform trajectory planning and check for collision.  However, performing tasks with 2D supervision is not desirable when the task is more complex, for example when manipulating objects in free space. If similar solutions were adopted, the user must master the ability to extrapolate from 2D images and reconstruct the 3D scene mentally. Such challenges have been commonly faced in surgery, where slices of 2D ultrasound or x-ray was used to perform complex procedures, such as in endovascular aneurysm repair \cite{doelare2021image}, albeit surgeons have undergone extensive specialist training in order to master that skill. On some occasions, the lack of information on image depth may even compromise patient safety \cite{ma2017augmented}. Overall, these challenges were commonly encountered in 2D workspace visualisation studies in literature, thus demonstrating the importance of a 3D visualisation experience.
With that said, 3D visualisation can take the form of full immersion (in the case of virtual reality, VR) and in the form of digital-reality fusion (in the case of augmented reality, AR). Under VR, the ability of the system to decouple the user from reality simplifies the problem of mapping from a user to a robot's reference frame. This translates the problem into the teleportation domain, where the user is effectively away from the scene of operation. However, this has led to problems such as poor user experience, difficulty in using the system, and inaccurate bi-directional mapping \cite{mine1997moving}. Lipton et al. has attempted a novel idea of using an intermediate ``VR control room" in order to give the user the same point of view as the robot, making the experience more realistic \cite{lipton2017baxter}. Nevertheless, VR technology is limited to teleoperation tasks, where the robot is instructed from afar about how it should move.
On the other hand, AR has gained popularity in the past decade, with its benefit of providing hands-free 3D guidance for a wide range of tasks. AR systems have matured over the past decade with the introduction of Microsoft HoloLens, an HMD system which capitalises on posture, gesture, vision, and voice to enrich the user experience, while also offering a vibrant AR developer's community that supports its fast growing user base. AR systems have been used in the past for applications in motion planning \cite{rosen2019communicating}, in control \cite{ostanin2019interactive}, and visualisation \cite{frank2017mobile}. Recently, AR has also been applied to the field of computer assisted neurosurgery, where mean target errors of 3.44mm on average \cite{kersten2015augmented} were achieved. The reader is directed towards the review by Makhataeva et al. \cite{makhataeva2020augmented} for more information of existing surgical systems. While enhancing the performance of precision medicine, AR systems have also received appraisal from their users. Rosen et al. performed a study that used an AR predictive system to inform the user about the fate of a trajectory and whether collisions will occur \cite{rosen2019communicating}. Data was collected from two user groups, with one being exposed to AR support and another without. At the end of the test, the accuracy of the two groups' predictions was evaluated, with the result indicating that the users of HMD yielded higher performance (75$\%$ vs 65$\%$). These studies demonstrate the pinnacle of state of the art AR systems and what they are capable of achieving.
In this work, we therefore explore the use of visual feedback for HRI in the form of workspace visualisation and reconfiguration guidance, applied specifically to reconfigurable robots.

\begin{figure}[t!]
    \centering
    \includegraphics[width=80mm]{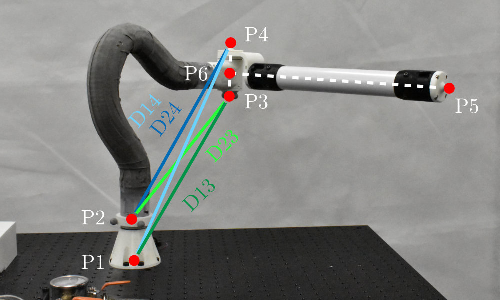}
    \caption{The previously developed 2-DOF malleable robot to be reconfigured, highlighting the point ($P1-5$) representation and interpoint distances ($D13$, $D14$, $D23$, $D24$) that define the malleable link topology.}
    \label{malleablerobot}
\end{figure}

\subsection{Augmented Reality System Design and Development}

To demystify the process of design and architectural building for the author, this subsection will begin with a brief introduction into the fundamental building blocks of an AR software, including, but not limited to, the Unity game engine, the C\# language, the elements of a scene that constitutes an AR experience, as well as the existing interaction strategies that the software engages with the user.

The AR application designed in this section was for a malleable robot system presented in~ Sect.~\ref{MalleableDesign}, and preliminary discussed in~\cite{clark2020design}. It is a 2-DOF reconfigurable robot, formed of a vertical revolute joint at the base, a malleable link connecting this base to a second revolute joint, and a rigid link connecting the second revolute joint to the end-effector. This is an extrinsic malleable robot, and thus we believe that there is potential for an AR system to recommend on the end-effector placement.

\begin{figure}[t!]
    \centering
    \scalebox{0.3}{
    \includegraphics{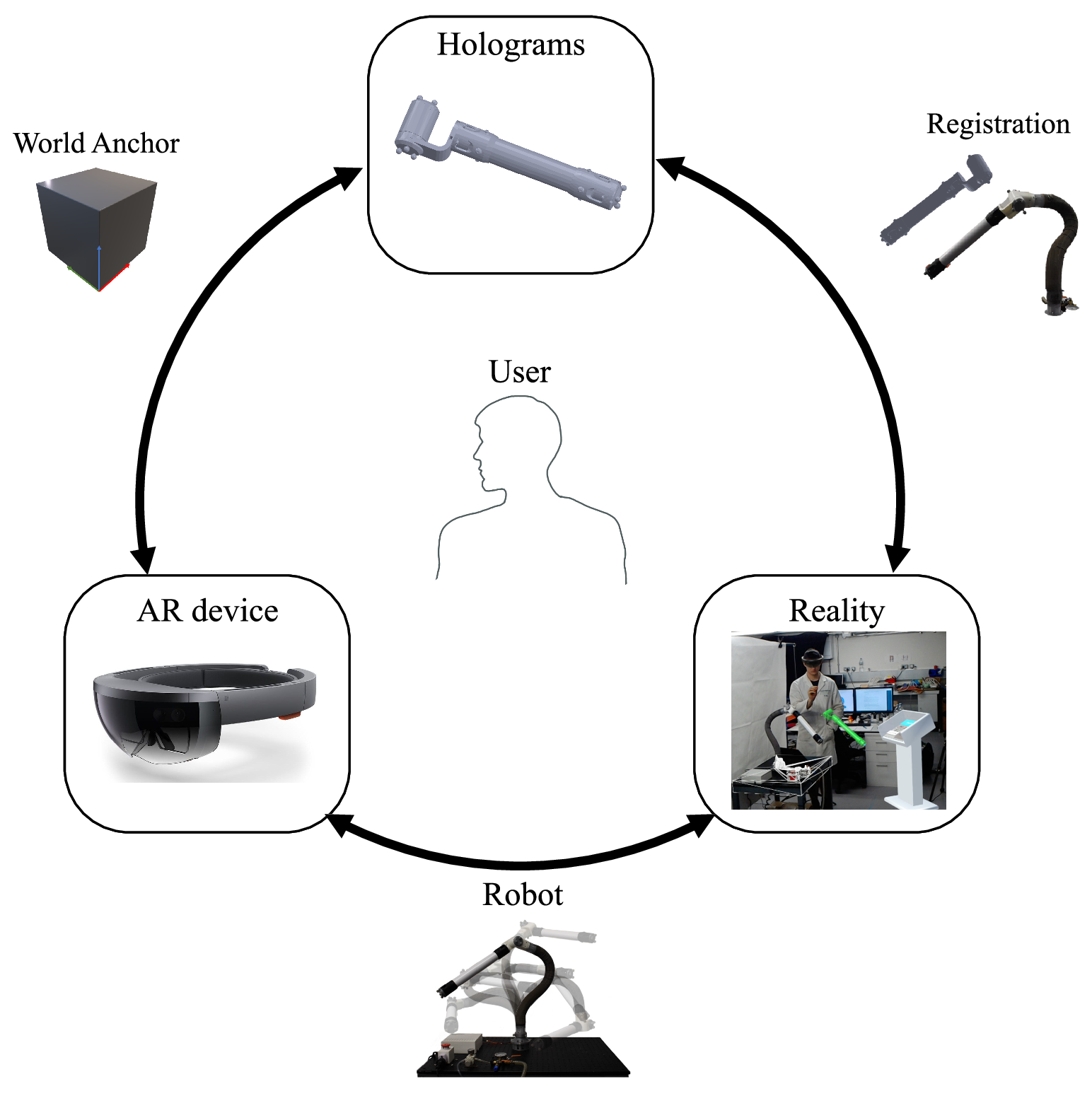}}
    \caption{Key AR components}
    \label{fig:Key_AR components}
\end{figure}

With any AR interface, there consists of three key components:
\begin{itemize}
    \item AR device (Microsoft Hololens): An HRI interface consisting of a display (head-mounted (HMD) or simply a device with a display) that facilitates user interactions. Such devices are not only capable of displaying virtual holographic objects, but also have the capability of receiving user intentions and actions through different modalities (Eg. Voice commands, gestures, device relative directions and positions...)
    \item Holograms (Unity game engine): These are objects that exist in the virtual world, but are displayed through the AR device. Such objects may take a shape which resembles the same object in the real world, or might be a fictitious one, eg. a virtual interactive panel.
    \item Reality (Malleable robot/Reconfigurable robot): This is the world that we as humans interact with on a day to day basis, and where the robotic system is located. Guidance from holograms and the virtual world should aid actions in the real world.
\end{itemize}
These components are highly intertwined with each other, and rely on each other's information in order to be defined or operate in harmony. For example, the definition of a hologram is dependent on the world anchor/origin, which is defined in the virtual world with respect to the AR device. When defining other holograms, the world anchor (or origin point) is used to define their positions. In order to make the position definitions as easy as possible, it is often convenient to define the world anchor with respect to a fixed object of interest, for example the base joint of a robot, as that is easy for the HMD as well as the user to see. The world anchor then needs to be registered to the real world, in order for the holograms to be defined at the correct locations. This can be done using optical tracking systems that consists of infrared cameras that outputs the absolute positions of fiducial markers with respect to its own origin, through markerless registration using surface point clouds or object landmarks with learning based approaches \cite{hu2021markerless}, or simply via manual alignment. Although the topic of registration is not the focus of this chapter, the reader may refer to \cite{andrews2020registration, makhataeva2020augmented} for further details of such algorithms. However, it should also be recognised that the importance of registration in the context of human robot interaction is debatable, as it cannot achieve the same degree of accuracy that a robot can. In addition, in the context of a malleable robot, the malleable link and the layer jamming mechanism limits the degree of accuracy achieved by the robot due to its flexibility and mechanical backlash.  Finally, there is the malleable robot that the user will manipulate based on the holographic guidance generated.
Having previously discussed the engineering design of the malleable robot, in the following section we focus on the main components within the virtual/holographic space, followed by a walkthrough of the main reconfiguration workflow.

\subsection{Holographic components}

\subsubsection{The table and the robot base}

\begin{figure}[!t]
    \centering
    \includegraphics{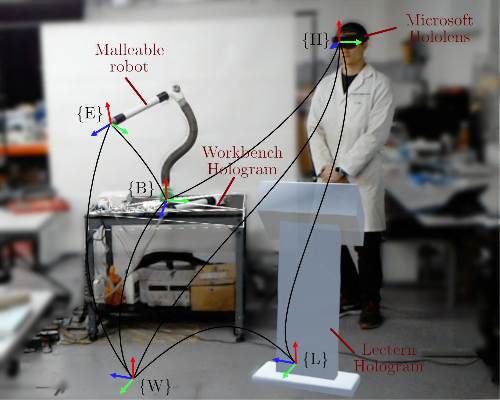}
    \caption{The Coordinate system of the scene, where \{H\},\{B\},\{L\},\{W\}, \{E\} represents the frames for the Hololens, the robot base, the lectern, the world origin and the end-effector, respectively}
    \label{fig:axes}
\end{figure}

In this application, the location and orientation definitions of the robot base is crucial, since all of the robot points are defined with respect to it (known also as the world anchor). The coordinate system of our scene can be represented by Fig.\ref{fig:axes}.

We denote the important frames of Hololens as {H}, the end-effector as {E}, the base as {B} and the world origin as {W}.  The Hololens, with its spatial awareness enables by its four tracking cameras, defines its spatial relation with respect to the robot base, and the world anchor., while the end effector is defined with respect to the robot base. The relative relations of the Lectern {L} is also seen by the Hololens, with its position defined by the user. 
Conveniently, the Unity game engine defines the relative relation between all objects and a world origin point \{W\}. In order to define everything with respect to \{B\} instead, a simple transformation needs to be done. For example, to find $\prescript{B}{E}{T}$, the transformation from the base to the end-effector, given the relationship between the end-effector $\prescript{E}{W}{T}$ with origin, and the base $\prescript{B}{W}{T}$ with the origin:
\small
\begin{align}
    \prescript{B}{E}{T} = \prescript{E}{W}{T^{-1}} \prescript{B}{W}{T}
\end{align}
\normalsize
Under Unity, this procedure can be accomplished by defining the local position of virtual objects with respect to the base frame \{B\} using a parent-child relationship.
We use the robot base to perform the initial alignment between the holographic world with the real world. To allow the user to deploy the system in any environment, we provide the user the freedom to drag and drop the holograms onto the real-world object, such that a manual alignment is performed. Automated solutions using fiducial markers such as QR codes, reflective markers or performing scene understanding are standard alternatives, but sending such information to the Hololens will inevitably increase the computation load, and we prioritized in reducing the load applied to the Hololens to perform such tasks.
In order to make the alignment more convenient, the robot base is attached to a larger object, the table, such that the user can manipulate the holograms more conveniently. We also added gesture controls to the object to allow for easy placement in free space, and voice commands to allows hand’s free “anchoring” and “releasing” of the world anchor. Furthermore, the table and robot base holograms are freely movable in position, but with its rotation in x and y fixed. This was done to improve the comfort of the alignment process, but more importantly it limits the alignment error and its effects on the system.
From a holistic standpoint, the robot base also constitutes a class of objects that is grouped under the robot base module. This consists of the table, the end-effector mesh models, the reconfiguration workspace, and the post-reconfiguration workspace. These objects will only be displayed during when conditions are triggered on the lectern, and when the task requires them.

\subsubsection{The lectern}
The lectern is the main interface that the user interacts with. It consists of a lectern hologram and a dynamic instruction panel on top. By pressing the buttons on the instruction panel, the user can navigate through the self-explanatory instructions, while the holographic scene changes with respect to the user proregression through the tasks. 
During initial development phases, the ergonomics of the lectern and a floating “tag-along” canvas was compared. The main difference between the two comes down to the preference of the user, however from preliminary user studies, we have found that having the instructions out of the way improves the workflow, as it does not accidentally obstructs the user’s view. In addition, the slanted design of the lecture allows the user to interact with it conveniently while they are standing up. In our case, it can be placed next to the working environment to take over the job of a traditional 2D display.
\subsubsection{The reconfiguration workspace} \label{The reconfiguration workspace}
Given the unique flexible nature of the malleable robot, there exists an infinite number of robot topologies (also known as reconfigurations) that can be achieved within a maximum reachable space \cite{clark2021malleable}. In order to narrow the reconfigurations down to a finite set of solutions, the system should respond to the specific requirements of the task, classify the nature of the task in terms of its feasibility through the use of distance geometry, then display this information back to the user with visual or other feedback mediums. 
In our application, the “reconfigurable workspace” (or otherwise known as the maximum reachable workspace) is defined as the volume in which all points within this space can be reached by the malleable robot through the tuning of stiffness of its link, such that there is no joint movement of the robot end-effector involved. The reconfigurable workspace is limited by the physical constraints of the robot, or more specifically, by the maximum reach ($L_{max}$)  and the minimum radius of curvature ($r_{min}$) of the malleable link, which corresponds to the involute and cycloid curves. This can be defined as follows:
\small
\begin{align}
    x &= r(1-(cos(t)+(t-a)sin(t))\\
    y &= r(sin(t)-(t-a)cos(t)),
\end{align}
\normalsize
and the cycloid curve is defined as:
\small
\begin{align}
    x &= r(1-cos(t))\\
    y &= r(t-sin(t))+L_{min},
\end{align}
\normalsize
where $a$ is the maximum bending angle of the malleable link, $t$ is the variable curvature of the malleable link, and $r$ is the radius of curvature. Values of $r_{min}=110$, $L_{max}=700$, $L_{min}=550$, and $a=5$~rad were used.
However, such curves still does not take into account the rigid distal link, which further extends this reconfigurable workspace. By offsetting this curve by the length of the distal link, then sweeping the 2D area enclosed by the involute and cycloid curves, we were able to generate the final reconfiguration volume for the interface.

\begin{figure}[t!]
    \centering
    \includegraphics[width=100mm]{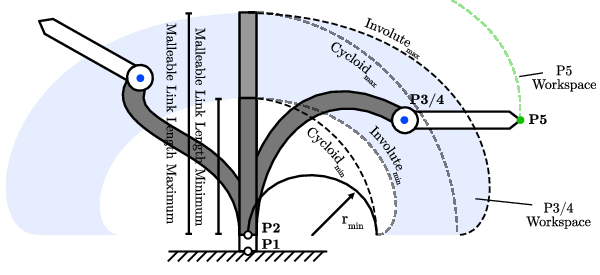}
    \caption{Involute and Cycloid curves traced by the malleable link in both maximum and minimum lengths, demonstrating how the reconfigurable space of P3/4 (\textbf{blue}) and P5 (\textbf{green}) were theoretically calculated.}
    \label{malleableLinkCurves}
\end{figure}

The reconfigurable workspace acts as a key decision-making tool for the user, as it directly represents all the possible reachable options that the robot can comply. Following several iterations of development and user feedback, the reconfigurable workspace was found to perform best when it is transparent and interactive, but constantly in the view of the user, as it can speed up the decision-making process. Additional checking algorithms were also implemented in the form of color feedback, such that the user can constantly verify the validity of their desired end-effector solution using collision detection algorithms from Unity.  When a valid solution is defined, the end-effector flashes green to indicate that the user can progress to the next step, or turns red if otherwise.

\subsubsection{The configured workspace and the choice of end effector solutions}
Following the selection of a target end effector pose, three tasks remain: the generation of a range of topologies which satisfies the user’s input, the selection of the suitable configuration that consists of a suitable workspace, the assistance of user alignment to desired configuration.
As discussed in section (insert Angus’ section on distance geometry), given the desired configuration, the topology computation equation computes a range of achievable solutions in the form of the 4 distances $D13$, $D14$, $D23$, and $D24$, which defines the distances between $P1$, $P2$, $P3$, and $P4$, corresponding to the robot base origin, directly above the origin, and the two sides of the distal joint respectively. Knowing those distances provides a deterministic pose of the end-effector. 
Visualising solely interpoint distances is difficult, and using them for reconfiguration is challenging even for experts in robotics. Nevertheless, in our previous work on end-effector alignment, a non-AR, tracking based alignment strategy was proposed [7]. An OptiTrack (Corvallis, Oregon, USA) system with 6 cameras was setup around the malleable robot. The user was given real-time feedback on the error between the expected position and the current position during an alignment, despite it being extremely difficult for the user to use such errors to correct their movement as there are no feedback regarding which direction this error is pointing towards. One alternative approach could be to use the AR interactive environment to mark the desired P3-5 positions in the user's vision, and physically attaching such points on the robot before performing the alignment. However, such system design will impact comfort and ease of use, since having only the 4 points means the user must deduce the position and orientation of the distal link.
To ease the user when performing alignment, we applied an additional level of abstraction by directly transforming the end effector using those 4 distances, such that the new robot topology is represented using the mesh model of the distal link, located at the final optimal position defined by the 4 distances. In order to do so, a simple transformation method was developed. Initially, the points $P3$ and $P4$ were defined using the distances, and $P3$ was used as the origin point $[0,0,0]$, where the z axis aligns to the line connecting $P3$ and $P4$. Then, we align the mesh of the distal link to $P3$, and rotate the model around $P3$ until $P4$ is aligned. For each combination of points $P3$ and $P4$, there exists a unique reconfigured workspace. Please note that this differs from the reconfiguration workspace, since it is the volume that the robot can reach given a particular end-effector reconfiguration. This workspace, is a subset of the larger reconfiguration workspace, defined when the malleable link is rigid, with the end effector placed at a unique pose.
During the rotation, multiple end-effector poses are generated, each possessing its unique workspace. Once generated, the mesh models of the end effector, and of the workspaces are loaded back into the AR scene to complete the reconfiguration. Using the interactive lectern, the user now loops through each of these poses, before selecting one that is most suitable for the task at hand. Finally, the user is guided with interactive holographic waypoints in the form of colored arrows, to depressurize the malleable link, and locking it in its desired place.

\subsubsection{Summary of workflow}

\begin{figure}[!t]
    \centering
    \scalebox{0.6}{
    \includegraphics{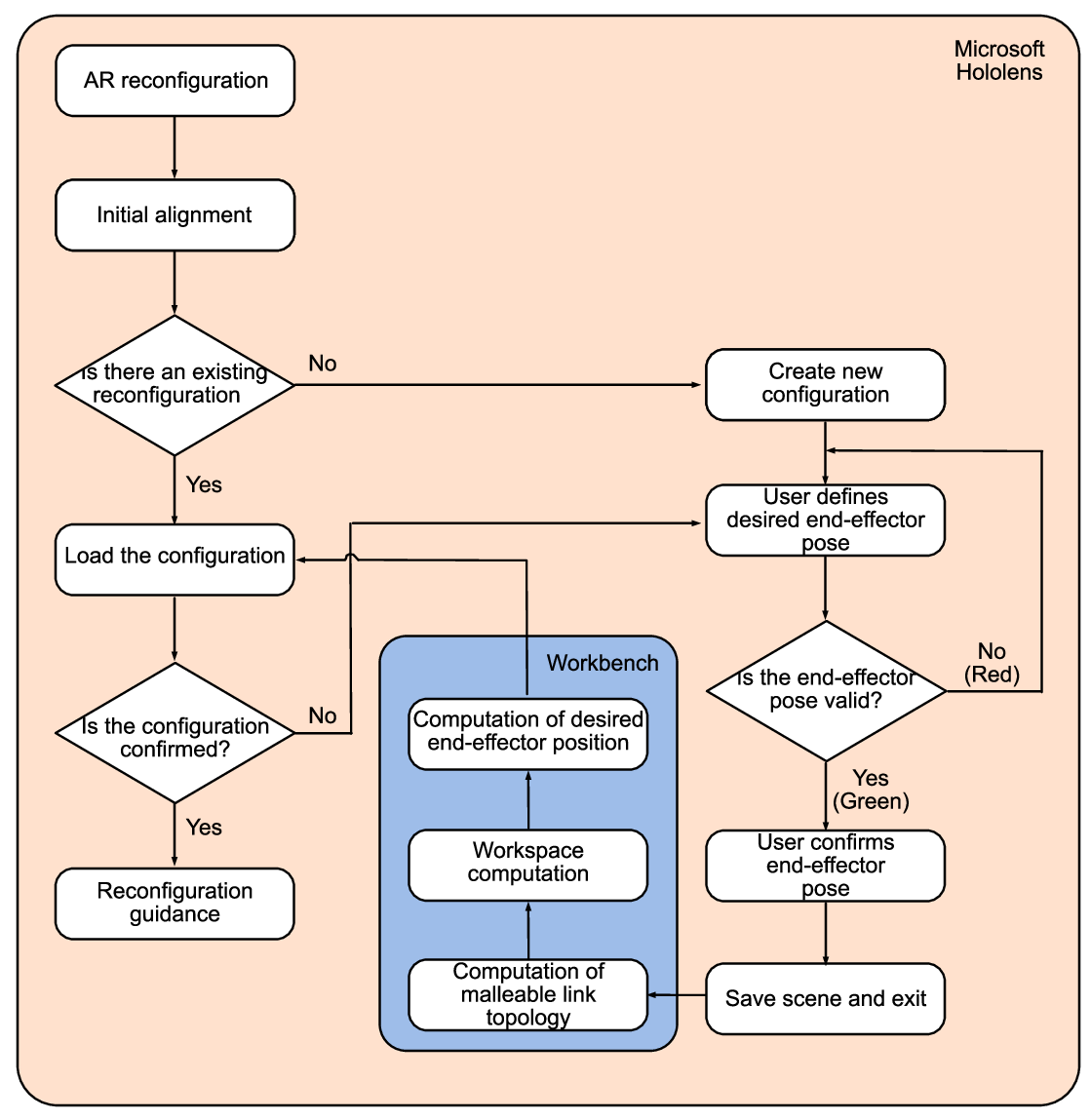}}
    \caption{A summary of the reconfiguration algorithm}
    \label{fig:Flowchart}
\end{figure}

The main ethos of the application is simplicity and ease of use, hence all of its elements and the instructions it provide are designed to be self-explanatory. Initially, the user performs an alignment between the real robot base and table with its holographic counter part, using voice commands such that they can adjust the manual alignment if the initial attempt was not successful. The user instructs the robot base to `anchor’ via voice command, which fixes the base in place. Thereafter, the user follows a standard procedure to perform the alignment in two distinct phases, definition and reconfiguration.
In definition, the user begins to configure a scene by observing the reconfiguration space from \ref{The reconfiguration workspace}, instantiating a mesh model of the end effector, then use hand gestures to orientate it to its desired pose. Once complete, the user confirms the position by pressing “check” on the lectern, where an end effector checking algorithm verifies the holograms lie within the reconfiguration space, and prevents invalid hologram transforms to be defined (eg. Outside the reconfiguration space). During checking, the end effector changes color to “green” indicating a valid topology, or “red” vice versa. Before the conditions for “check” is fulfilled, the user is prevented from advancing further, providing a fail safe.
At this stage, the user then confirms the coordinates of the distal link points P5 and P6, then proceed to saving the configuration in the scene, and send these information to the workstation for computation. The workstation is capable of computing the optimal topologies of the malleable robot, defined by distance geometry using a sampling of dihedral angles between the triangles formed by the point in question (P4/5/6), the orientation of the end effector and the fixed robot topology, and returns the desired end effector poses, along with their workspaces (as discussed in Section x). The offloading of the mathematical computation allows the Hololens to perform at high frame rates, needed for real time guidance. 
The final results from the workstation is now transferred to the Hololens. In this implementation, such transfer is performed by saving the scene, closing the application and redeploying to the Hololens. The obvious disadvantage of doing so is the disruption made to the workflow, and the interruption introduced during redeployment. On the other hand, TCP/IP communication is a clear alternative, where a seamless transition between the two phases of workflow is ensured. However, this decision was made in consideration of the hardware limitations of the Hololens, where its computation power is simply not sufficient to handle a mass transfer of mesh models on the fly, without substantially affecting the user experience (eg. Freezing, very low fps rate). In future iterations of this project, and with the acquisition of better head mounted display devices, researchers will be able to incorporate TCP/IP into the software without significant performance.

\section{Autonomous Reconfiguration}

\subsection{Introduction to Control of Flexible Spines}

Malleable robot design incorporates a flexible spine and layer jamming components. The author first investigated controlling the flexible spine of malleable robots using the tendon-driven approach. Unlike the traditional continuum robot with a centre spine, the flexible spine has four soft spines spaced equally around the disk and is compressible. The advantage of using flexible spines is that it allows cables to pass through the center of the malleable robot. 

Traditional control of continuum robots mainly relies on an accurate mathematical model. Beam theory, Cosserat rod model and Constant curvature model are commonly used in modelling the continuum robot. Cosserat rod model is the most accurate among all the continuum robot models with consideration of external loads while being computationally expensive \cite{rao2021model} \cite{till2019real}. Alternatively, a learning approaching using neural network to model system of soft continuum robots is popular due to non-linear characteristics of soft materials. Machine learning approaches like gaussian process regression, Kalman filter, and Koopman operator model have been used in controlling continuum robots trained with real-world data  \cite{fang2019vision}\cite{deisenroth2011pilco}\cite{sabelhaus2021gaussian}\cite{li2017model}. Deep learning methods are divided into model-free and model-based learning methods. Model-free methods like deep Q-learning utilize neural network and gradient descent to find optimal policy to control continuum robots. Model-based learning methods have also been proposed, mainly relying on a math model. When these deep learning methods trained in simulation, there is a sim-to-real gap needed to fill. Meanwhile, training them in the real-world is not data-efficient. 

This section focuses on solving the problem of data-hungry in learning-based methods for controlling continuum robots. A real-world data-efficient model-based learning framework has been proposed, requiring only 100 real-world data to achieve comparable or better performance compared to control policies trained with thousands of real data. 
\begin{figure*}[ht!]
 \centering
 \includegraphics[width=\textwidth]{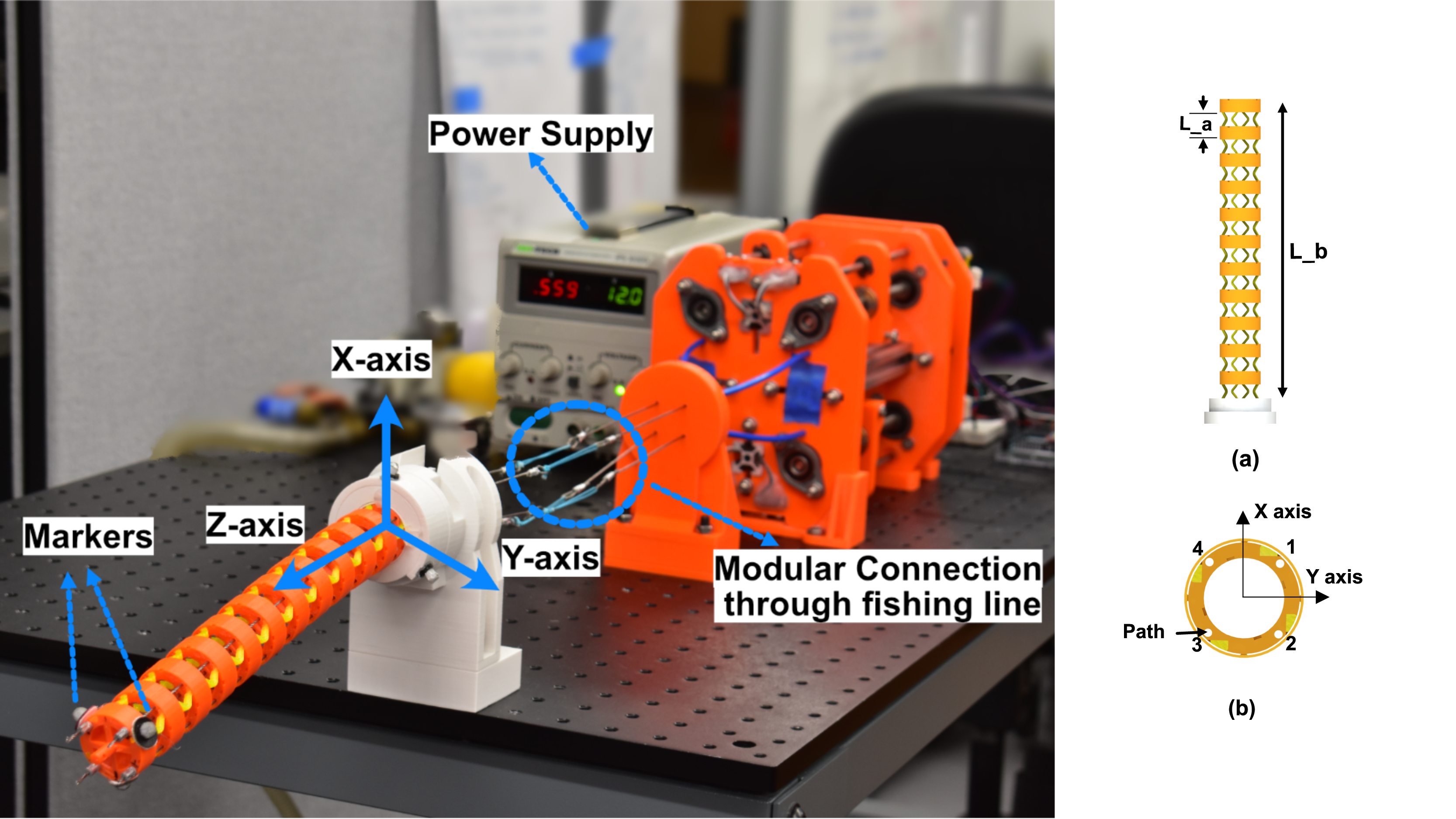} 
 \caption{(a) Design of flexible spine.(b)Tendon path and cross-section area in flexible spine}
 \label{fig:X1}
\end{figure*}

\subsubsection{Flexible Spine Design and Continuum Robot}

The continuum robot we experimented with has four compressible spines. The spine is 3D-printed using thermoplastic polyurethane (TPU). The disks are 3D-printed using PETG material. We connect the four spines with 12 plastic disks with 0.01 m space between them in the relaxed state. The overall length of the continuum robot is 0.22 m. This type of continuum robot is easy to manufacture, and inherently safe to work with due to its flexibility. There are four cables passing through the continuum robot, each controlled by a stepper motor. Meanwhile, the position control is performed by adding position switches to reset each time the robot is powered on. 

As shown in \ref{fig:X1}(b), a unique connector was created to join tendons and a lead screw (b). A cable tensioner was included into the 3D-printed connector. By pushing the cable seat with M4 bolts and nuts in the connector, the total length of the cable is shortened. Tendon tension is increased, and the cable is held in place. The incompressible Teflon tubes serve as the continuum robot's guiding routes for the steel-wire (tendons). 

\subsection{Mathematical Modelling through Cosserat Rod Model}
The continuum robot with soft spine is easy to deform under external loads. In this case, cosserat rod model provides us the advantage of accurate modelling while we are able to adjust the model to capture the effect of compression that reduces the overall length of the continuum robot. 

The idea of cosserat rod model is to divide the continuum robot into finite small sections and apply force and moment balancing equation along these small sections (can reference chapter 2).  With known boundary value conditions, we can use the shooting method to solve such problem. In this case, our continuum robot is fixed at one end. And on the other end, assuming we have known its position and orientation to simplify the problem, this becomes a boundary value problem we can solve by having different initial guesses of moments and forces acting on the continuum robot. 

When the robot is tendon actuated, the system's given tendon lengths limit the position and orientation of the distal end of the continuum robot.  In order to convert this into a new boundary condition, we must integrate the tendon route length to match the given tendon length while making guesses about the moments and forces acting on the continuum robot using the shooting approach.

The Cosserat rod model was first developed in the static model based on the kinematics of a rod. Along the centre-line of the rod, we have defined the position $p(s)$, orientation $R(s)$, the change rate of linear velocity $v(s)$ and angular velocity $u(s)$ at the length s of the rod. For any arbitrary section of a rod,  static equilibrium equations of force and moment are applied. Using linear constitutive equations to map internal force $n(s)$ and moment $m(s)$ of a rod, the explicit Cosserat static rod model can be represented as:
 \begin{align}
     \begin{split}
          \dot{p} & = Rv, v = K_{se}^{-1}R^Tn + v^{*}\\
          \dot{R} & = R\widehat{u}, u = K_{bt}^{-1}R^Tm + u^{*}\\
          \dot{n} & = -f\\
          \dot{m} & = -\dot{p}\times{n}-l
      \end{split}
      \label{eq:X1}
 \end{align}
where $v^*$ and $u^*$ are reference configurations of a straight rod, in our case, $v^*$ is [0;0;1] and $u^*$ is [0;0;0]. And $K_{se}$ is the stiffness matrix for shear and extension and $K_{bt}$ is the stiffness matrix for bending and torsion. They can be represented as:

\begin{align}
    \begin{split}
        K_{se}(s) & =diag(GA(s), GA(s), EA(s))\\
        K_{bt}(s) & =diag(EI_{xx}(s), EI_{yy}(s)), EI_{zz}(s))\\
    \end{split}
    \label{eq:X2}
\end{align}
where $G(s)$ is the shear modulus, $A(s)$ is the cross-section area of the rod, $E(s)$ is Young's modulus, $I_{xx}$ and $I_{yy}$ are the second moments of the in the corresponding axis. $I_{yy}=I_{xx}+I_{yy}$. 

\subsubsection{Tendon model}
 The tendon path is defined as radius relative to the centre of the spine $r_{i}(s) =[x_{i}(s)\: y_{i}(s)\: 0]^{T}$ knowing the $x$,$y$ coordinates of each tendon, assuming the longest dimension of continuum robot inline with $z$ axis. For example, in figure \ref{fig:X1}, the cross-section of the continuum robot stays constant along the length $s$ of the continuum robot. $r_{i}$ becomes fixed value across the continuum robot. Otherwise, it is a function of length $s$ of the continuum robot. The tendon position relative to the centre of the spine with displacement using previous radius and current orientaion of the rod can be represented as $p_{i}(s)  =R(s)r_{i}(s)+p(s)$, 
 
In this tendon force model, we assumed no friction between the tendon and the path it travels to simplify the model. Using static equilibrium equations in Equation \eqref{eq:X1}, we represent the internal force and moment distribution caused by tendons as following:
\begin{align}
   \begin{split}
       f_{t}=-\sum_{i=1}^{n}\tau_{i}\frac{\hat{\dot{p}}^{2}_{i}}{\left \|\dot{p}_{i}  \right \|^{3}}\ddot{p}_{i},\, 
        l_{t}=-\sum_{i=1}^{n}\tau_{i}\hat{(Rr_{i})}\frac{\hat{\dot{p}}^{2}_{i}}{\left \|\dot{p}_{i}  \right \|^{3}}\ddot{p}_{i}\\
     \end{split}
     \label{eq:X3}
\end{align}

\subsubsection{Tendon-driven Cosserat Model}
Now we incorporate the tendon model in $f_{t}$ and $l_{t}$ with a static Cosserat rod model. The derived tendon-control static model can be represented as:

\begin{align}
    \begin{split}
    \dot{p} & = Rv, v = K_{se}^{-1}R^Tn + v^{*}\\
    \dot{R} & = R\widehat{u}, u = K_{bt}^{-1}R^Tm + u^{*}\\
    \dot{n} & = -f_t -\bar{f}\\
    \dot{m} & = -\dot{p}\times{n}-l_t-\bar{l}\\
     \end{split}
     \label{eq:X4}
\end{align}

where, $-f_t$ represents distributed tendon tension forces, $-l_t$ represents distributed moment caused by tendon forces. In addition, $-\bar{f}$ and $-\bar{l}$ represent the distributed forces and moments caused by the continuum robot and external loading. 

\subsubsection{Boundary Conditions for Tendon-length Control using Static Cosserat Model}

When comes to using tendon length control, we now guess tendon forces in each tendon and reduce the tendon length error in Error term for given tendon length value for each tendon using a shooting method. The boundary conditions are described as the moment and forces at the tip of the continuum of robots. We first have an initial guess on $v(t,0), u(t,0)$, which are the velocity and angular velocity for $\forall$ time $t$ and at length 0 of the rod. Then we applied the shooting method to reduce residual terms for given tendon forces $\tau_i$ in each tendon The shooting method is solved in MATLAB software's fsolve function. Now the initial guess and residual term represented as following:

\begin{align}
    \begin{split}
        Guess &= [v(t,0)^T \ u(t,0)^T \ \tau_1(t) \ ... \ T_n(t)]\\
        Error &= [(E^F)^T \ (E^M)^T \ E^l_1 \ ... \ E^l_n]\\
        E^l_n &= (1+\tau_i)(l^*-l_i+\beta_i)\\
        l_i &= l^q_i+l^B_i\\
        \beta_i &= \begin{Bmatrix}
        0, & \tau_i > 0\\ 
        \gamma^2,  & \tau_i < 0
        \end{Bmatrix}\\
        \tau_i &= \begin{Bmatrix}
        \gamma^2, & \tau_i > 0\\ 
        0,  & \tau_i < 0
        \end{Bmatrix}\\
     \end{split}
     \label{eq:X5}
\end{align}

where $l^q_i$ is a fixed length from the end of the continuum robots to the motor, $l^B_i$ is an integrated length along the backbone of the continuum robots, and the $\beta_i$ is a slack that existed in each tendon. 

The slack is defined as if the tendon force $\tau_i$ is great than zero, then there will be no slack in the tendon. Otherwise, it will have a positive amount of tendon slack, which lets the shooting method optimize other tendons with positive tendon forces.

It is worth noticing that the residual terms need to change according to the number of actuating tendons. If there is only one tendon actuating, only $E^{l}_{1}$ is needed to include in the Error term along with the initial guess of the tendon force $\tau_1$.
Otherwise, it is restricting three other tendons from exceeding the original length of the continuum robots. If two tendons are actuating the continuum robot, $E^l_{1}$, $E^l_{2}$ and initial guess of $\tau_1$, $\tau_2$ are included corresponding to those two tendons and so on.

\subsubsection{Compressible Spine Modelling}

Tendon tensions, which are compression forces, account for the majority of internal forces in the Cosserat rod model. We have taken into consideration the internal forces that compress the spine under the circumstance of low external stress and the continuum robot's lightweight. In this case, we have assumed that the robot is under compression mostly from internal forces generated by tendon tensions. 

To characterize the spine compression under tendon tension, we have conducted multiple trials of experiments to determine the length change of our compressible soft spine with respect to the compression force. Firstly, we used a two-section specimen of our compressible spine for testing. A simplified linear equation has been used to represent the relationship between our specimen's changing length and the various force applied on top of it:
\begin{align}
    \begin{split}
        l_c&=c_{spine}\times{\left \|  n(s)\right \|},\\
     \end{split}
     \label{eq:X6}
\end{align}
where $\left \|  n(s)\right \|$ represents the magnitude of internal normal forces acting on the robot. The experiments conclude the coefficient has a value of $0.2\frac{mm}{N}$ with a limit compression force of 20 Newtons.

As illustrated in Figure \ref{fig:X1}(b), our continuum robot consists of eleven sections of compressible segments. The weight and external load effects on the interior forces $n(s)$ at these 11 sections are calculated using linear constitutive law:

\begin{align}
    \begin{split}
        n(s) & =R[K_{se}(v-v^*)],   
    \end{split}
    \label{eq:X7}
\end{align}

where $K_{se}$ is the stiffness matrix for shear and extension and $v^*$ is the straight rod reference configuration with a value of [0;0;1].
Instead of simply reducing the total length of the continuum robots, we iteratively reduced the length of each segment of the continuum robots according to the internal force at the length s of the rod within the shooting method of the Cosserat rod model. The algorithm is described in detail in Algorithm 1. The shooting method continues to optimize the function using MATLAB function fsolve until it reaches a desired error threshold $\epsilon_t$. We discretize the continuum robot into many small sections, $ds(i)$ represents the section length, which is the total length  of the continuum robot divided by the number of sections. 

\begin{algorithm}[t]
\caption{Calculating  the reduced length of the continuum robot with compressible spine}
\label{alg:X1}
\begin{algorithmic}
\While{Error is greater than  $\epsilon_t$}
\State Solving system ODEs for $\dot{p}, \dot{R}, \dot{n}, \dot{m}$
\State Using Euler's method to calculate $p,R,n,m$
\ForEach{ compressible section i, i = 1,2,...,11}
\ForEach{ Spatial Point j inside section i}
\State Calculate $n(i)$ at segment $i$ with Equation \eqref{eq:X6}
\State Calculate $l_c(i)$ using Equation \eqref{eq:X6}
\State Update $ds(i)$ as $ds(i)=ds(i)-l_c(i)$
\State Update $p(j),R(j),n(j),m(j)$ using Euler's Integration method
\EndFor
\EndFor
\EndWhile
\end{algorithmic}
\end{algorithm}

\subsection{Neural Network and Gaussian Process}

We introduce a model-based learning method by training a Cosserat rod mathmatical model using a neural network. A Gaussian process regression (GPR) to model the error between the simulated model and real model. This error model will later be incorporated into the control algorithm to achieve data-efficient control. Three control policies based on the GPR error model and a neural network trained in the simulation are formed to provide a real-world data-efficient framework to control the continuum robot.

\subsubsection{Neural network modelling for the System}
As a universal approximator, neural networks have been extensively used to model robotic systems. In terms of continuous robot control, neural networks have outperformed many computationally costly methods, such as finite elements and finite differences, in terms of computational efficiency. In both kinematic and dynamic modeling of continuum robotics, neural network-based approaches have demonstrated comparable or lower error prediction than traditional mathematical models.

\begin{figure*}[t!]
 \centering
 \includegraphics[width=\textwidth]{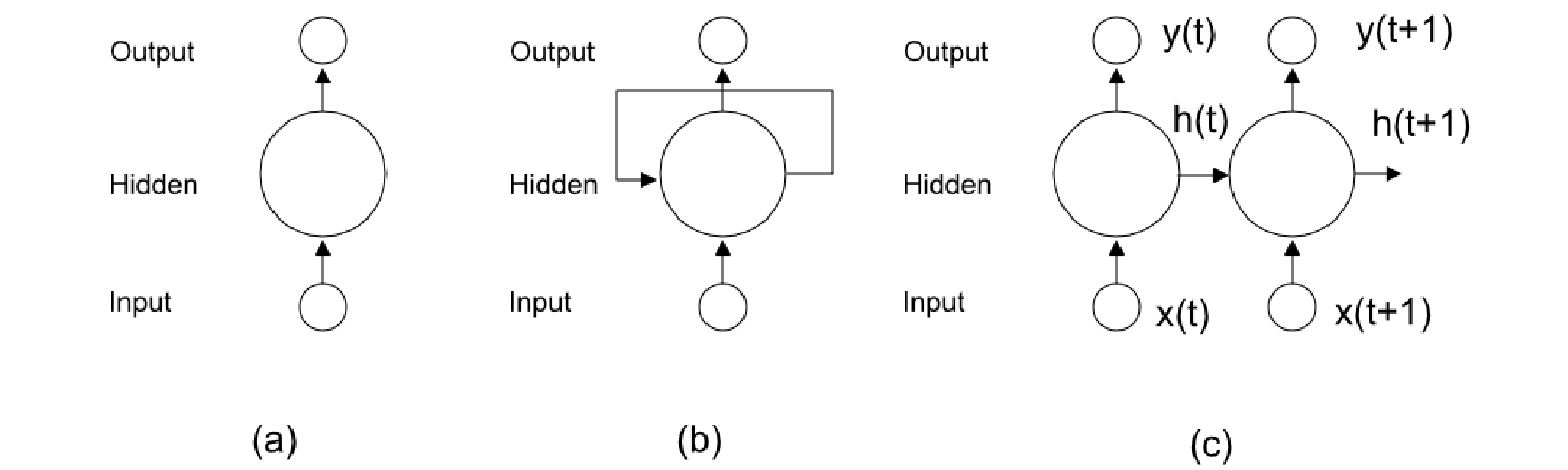} 
 \caption{(a) Feedforward neural network.(b) Recurrent neural network.(c) Unfolded Recurrent neural network.}
 \label{fig:X2}
\end{figure*}

To use neural network to simulate such continuum robot system with soft spines, a feed-forward neural network or recurrent neural network was typically used. By providing the control inputs of the systems and resulted end effector positions, the neural network could predict the positions of the continuum robot system after training. The difference between a feed forward neural network and a recurrent neural network is presented in the Figure \ref{fig:X2}. The benefits of recurrent neural network are to incorporate history into the training, which allows for better capture of dynamic systems. To simulate a static continuum robot, a feed-forward network is typically used in this case. 

For modeling the continuum robot with a feed-forward neural network, there are few observable variables in the system. The continuum robot is controlled through four tendons $L_1, L_2, L_3, L_4$ through the spine, while the end effector position $p_i$ in $X,Y,Z$ axises of the continuum robot is captured using the motion tracking system. By providing the control inputs parameters and corresponding end-effector positions, we can specify the system like this:

\begin{align}
    \begin{split}
        &u_i=(L_1,L_2,L_3,L_4)\\
        &(u_i,p_i)\rightarrow p_{i+1}\\
    \end{split}
    \label{eq:X8}
\end{align}

For each given control input, and the current position of the end effector, we can predict the resulting end-effector position using a neural network. The structure of the neural network can be constructed as an input layer, hidden layers, and an output layer.  Taking feed-forward neural network as an example, the hidden layers can be represented as Dense layers using Keras API. In the case of recurrent neural network, the hidden layers is easily represented using simple RNN layers or LSTM (Long Short Term Memory networks) layers.  

\subsubsection{Data Generation for Neural Network Training}
\label{xin_neural_network_section}

To train such neural networks, we first need to generate data from Cosserat rod mathematical model. Given different control inputs $u_i$ and initial guess of velocity $v(t, 0)$, and angular velocity $u(t, 0)$ for $\forall$ time $t$ and at length 0 of the rod, we can use shooting method to estimate the end-effector position of the continuum robot. 

In the calibration process of the mathematical model, we use a small number of real position data and corresponding control inputs to adjust Young's modulus (E) of the Cosserat rod model. To perform this optimization process that minimizes the error of simulated end-effector positions and real position data, we can use MATLAB function:fminsearch. 

A random motor bubbling technique is used during this data generation process. This method is to add random offsets to the control inputs and generates the corresponding mathematical position data points. By generating data this way, it helps to better capture the system and the trained neural network is less likely to overfit. 10000 data points have generated in this way for neural network training. The training of the neural network is straightforward with a choice of "adam" optimizer, "mean squared error" in the Keras sequential model, and the option of "EarlyStopping" to avoid overfitting. 

Since this continuum robot has a soft spine, the calibrated mathematical model alone still has relatively large errors compared to the traditional continuum robot with a rigid spine. The same is true for neural networks trained from this calibrated mathematical model. In the following section, we present methods that incorporate Gaussian process regression with our trained neural network  to achieve accurate modelling while being data-efficient. 

\subsubsection{Gaussian Process Regression}
Gaussian process regression has been demonstrated to be data-efficient in model-based system control with fewer data points \cite{deisenroth2011pilco}. It can provide continuous mean value and covariance value estimation for unknown data points using  observed data points. The Gaussian process has two main components: a mean function  and a Kennel function / Covariance function. Normally, the mean value is constant and assumed to be zero. The covariance function encodes information regarding the expected shape and structure of the function. A simple and popular example is the square exponential (SE) kernel. This also implies if two points in the input space are close to each other, then the outputs (or function values) at those points are likely to be similar as well. GPR could approximate any nonlinear function $y_i\sim f(x_i) + \epsilon$, where $\epsilon$ is a white Gaussian noise with zero mean and variance $\sigma^2_n$. The GBP and SE kernel is represented as:
\begin{align}
    \begin{split}
        &f(x_i) \sim GP(m(x_i), k(x_i,x'))\\
        &k(x_i,x') = \sigma^2_sexp(-\frac{1}{2}(x_i-x')^T\lambda(x_i-x'))\\
    \end{split}
    \label{eq:X9}
\end{align}

Now we consider an input data $X={x_i}$ and output data $Y={y_i}$ for i = 1,2,...,N. We can select a GP prior distribution on $\mathbf{f}$ for the function $\mathbf{f}=f(X)$ as follows:

\begin{align}
    \begin{split}
        p(\mathbf{f} | X) = N (\mathbf{f}; \mu(X), K(X, X))
    \end{split}
    \label{eq:X10}
\end{align}

Now given some observed data points $D=(X,\mathbf{f})$, we want to make predictions on points $X_*$. We can write the joint distribution between the known training values $f(X)=\mathbf{f}$ and testing function values $f(X_*)=\mathbf{f}_*$ as follows:

\begin{align}
    \begin{split}
    p(\mathbf{f},\mathbf{f_*}) = N 
    \big(
      \begin{bmatrix}
        f \\ f_*
      \end{bmatrix};
      \begin{bmatrix}
        µ(X) \\ µ(X_*)
      \end{bmatrix},
      \begin{bmatrix}
        K(X, X)+\sigma^2_nI &  K(X, X_*) \\ K(X_*, X) & K(X_*, X_*)
      \end{bmatrix}
      \big)
    \end{split}
    \label{eq:X11}
\end{align}

Then the posterior distribution to make predictions for input set $x_*$ can be expressed as follows:

\begin{align}
    p(\mathbf{f}_* | X_*, D) &= N(\mathbf{f_*}; m_{f|D}(X_*), K_{f|D}(X_*, X_*))\\
             m_{f|D}(x) &= m(x) + K(x, X)(K(X,X)+\sigma^2_nI)^{-1}(y - m(X))\\
             K_{f|D}(x,x') &= K(x, x') - K(x, X)(K(X,X)+\sigma^2_nI)^{-1}K(X, x')
\label{eq:19}
\end{align}

There is also a drawback, for a more complex system, more data points will be needed in the training process. While GPR becomes more computation expensive with larger amount of training data points. To overcome this drawback, we have selected GPR to model the error term between the data generated from Cosserat rod model and real experiemtn data. The detailed  inputs and outputs of our GPR model is described in the following equation:

\begin{align}
    \begin{split}
        x_i&=(L_1,L_2,L_3,L_4, p_{sim})\\
        f(x_i) &= p_{sim}-p_{exp} = e_i\\
    \end{split}
    \label{eq:X12}
\end{align}

Where $p_{sim}$ represents the end-effector position given control inputs of $L_1,L_2,L_3,L_4$ from the Cosserat rod model. In this way, we could dramatically reduce the number of real data points to only 100 real data points to get an accurate GPR model as explained in \cite{wang2022data}. By providing observed data points, the GPR now is ready to make predictions on new data points. The detailed implementation of a Jupiter notebook can also be found on our remote repository. 

\subsection{Data efficient Control Policies}

\subsubsection{Inverse Mapping of the Continuum Robot}

In the section \ref{xin_neural_network_section}, we used a neural network to model the forward kinematics of the continuum robot system. Now in order to control the continuum robot to reach certain positions, we need an inverse mapping of the continuum robot. In the motor bubbling data generation stage, we recorded, control input $u_i$,current position $p_i$ and next position $p_{i+1}$. In this case, we just need to rearrange the inputs and train the neural network. The structure of the neural network is the same. 

\begin{align}
    \begin{split}
        (p_i, p_{i+1})\rightarrow u_i\\
    \end{split}
    \label{eq:X13}
\end{align}

To apply the trained inverse mapping neural network, we have a current position and desired position, then we could easily obtain corresponding control input in this case. The representation can be written like this:

\begin{align}
    \begin{split}
        (p_i, p_{target})\rightarrow u_i\\
    \end{split}
    \label{eq:X14}
\end{align}


\begin{figure*}[t]
 \centering
 \includegraphics[width=\textwidth]{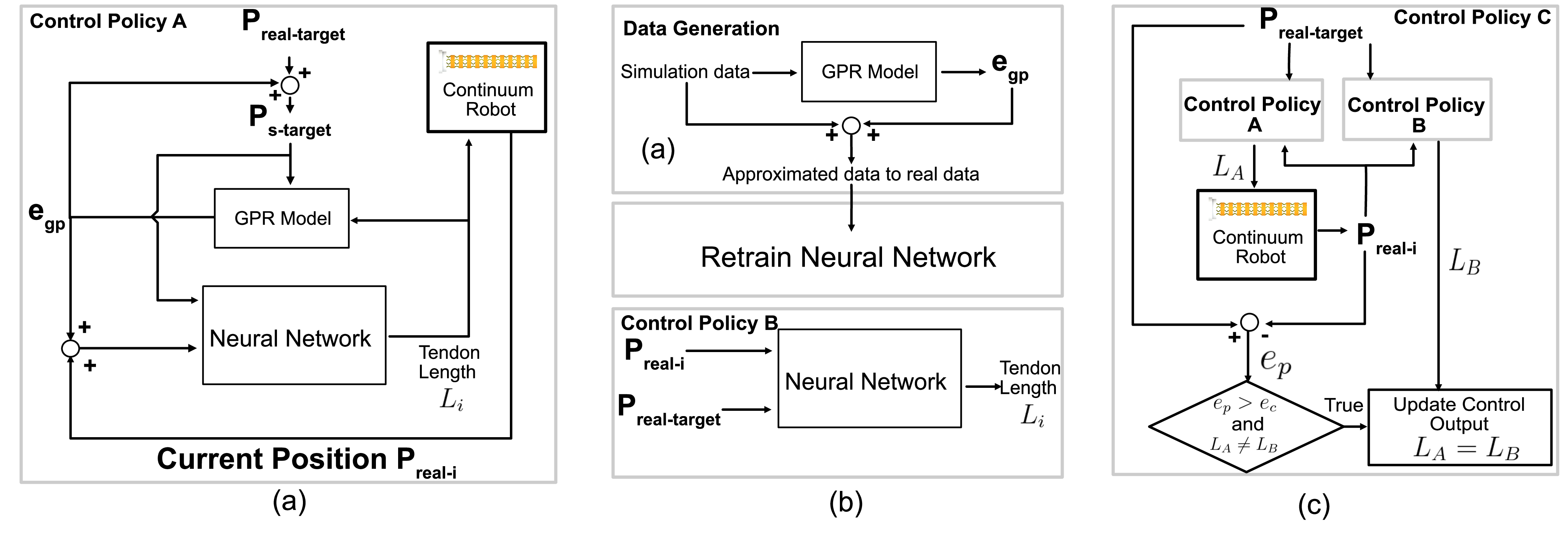} 
 \caption{(a) Control Policy A.(b) Control Policy B.(c) Control Policy C.}
 \label{fig:X3}
\end{figure*}

\subsubsection{Control Policy A}

The inverse mapping of the continuum robot is trained using simulation data, but there is still a sim-to-real gap. The neural network trained directly from the mathematical model still has noticeable errors. GPR error model in this case can be used to fill the sim-to-real gap. 

\begin{align}
    \begin{split}
        e_{gp} = f((L, p_{s-target}))\\
        p_{s-target} = p_{real\: target} + e_{gp}\\
    \end{split}
    \label{eq:X15}
\end{align}

Given a target position $p_{real\: target}$ for the continuum robot to reach in the real-world space, we want to find the corresponding control inputs. We have an inverse mapping neural network trained from mathematical equations(simulation). However, in order to allow this neural network to generate correct control output. This $p_{real\: target}$ needs to be compensated with an error term $e_{gp}$, such that we could obtain a simulated target position $p_{s-target}$. 

Similarly, the error term $e_{gp}$ also depends on this simulated target position. This will need an iterative approach that continuously updates this $p_{s-target}$ value. The initial guess, we can assume  $p_{s-target} = p_{real\:  target}$. Later, this value will be updated as the continuum robot move close to the target position. 

When given a real target point $p_{real\: target}$ in the real workspace, now we need to estimate the target point $p_{s-target}$ in the simulated workspace(. Term $e_{gp}$ is also dependent on the input value of  $p_{s-target}$. It then becomes an optimization problem of finding suitable $p_{s-target}$ value. We take an initial guess of $p_{s-target} = p_{real\:  target}$. feeding it into the RNN and receive tendon length control outputs. Then feeding the control outputs and $p_{s-target}$ into the GPR model to generate $e_{gp}$. This error term will be added to $p_{s-target}$ as a correction of initial guess.

For open-loop control, the current position remains unchanged during the control time step. In contrast, for closed-loop control, the current position is continuously updated based on position feedback. In the meantime, as an input for our inverse mapping model, we assume the current position is from the simulated workspace rather than the actual workspace. The current position will also need to be compensated with an error term  $e_gp$ from the GPR model as well. As a result, the error term becomes accurate as the robot reaches the target point in the closed-loop control. The controller is presented in Algorithm 2 and Fig \ref{fig:X3}.

\begin{algorithm}[t]
\caption{Update $p_{s-target}$ with iterative approach in closed-loop control}
\label{alg:X2}
\begin{algorithmic}
\Require $p_{real\: target}, p_{inital}$, time-step: $i$
\State initial guess: $p_{s-target} = p_{real\:target}$
\While{$\left \| p_{s-target}(i-1) - p_{s-target}(i) \right \| > 0.001 $}

\State Get control input  $(p_{initial},p_{s-target}(i-1))\rightarrow L(i)$
\State Predict using GPR $e_{gp}(i)=f((L(i),p_{s-target}(i-1)))$
\State Obtain tip position $p_{real(i)}$ from  tracking system
\State Update current position $p_{inital} = p_{real}(i) +  e_{gp}(i)$
\State $p_{s-target}(i) = p_{real\: target}+e_{gp}(i) $

\EndWhile
\end{algorithmic}
\end{algorithm}

\subsubsection{Control Policy B}

 The GPR error model could also be used to generate more data given simulated inputs as the GPR model could predict continuously mean and covariance values for different unseen inputs. In this way, we using Equation \eqref{eq:X12} to generate corresponding error term $e_{gp}$. Assuming that each $e_gp$ is accurate, we will obtain accurate real workspace positions by utilizing the following equation to approximate real position data for given control inputs:
\begin{align}
    \begin{split}
    p_{exp} = p_{sim}-e_{gp}\\
    (p_{real-i}, p_{real-target})\rightarrow u_i\\
    \end{split}
    \label{eq:X16}
\end{align}

With updated training data, we can now retrain the inverse mapping neural network. This method is more straightforward, the four tendon lengths are directly obtained from the retrained neural network given target position in the real workspace. However, control policy A has slightly better performance based on experiment data \cite{wang2022data}.

\subsubsection{Control Policy C}

This control policy incorporates elements of control policies A and B to create a hybrid model. The A control policy relies on a precise simulation model, while the GPR model compensates for the error. Typically, the simulation model is a good approximation of the actual robot, and when it is accurate, it performs well.

Due to the simulation model's simplifications and numerous assumptions, certain physics could not be accurately represented. This issue becomes more critical for our application as we incorporate softer materials TPU. Using the same data generated by the GPR model, we employ control policy B to learn the mechanics that are absent or severely distorted in the actual robot. Although the approximated data generated by the GPR model may be less precise than actual data, it will outperform control policy A when the simulation model is inaccurate. 

Using the final position tracking error as a criterion and checking $L_A$ and $L_B$ to prevent an infinite cycle, we have presented the control policy C to connect policies A and B. Presented in Figure \ref{fig:X3}, if the absolute position error is greater than a constant value $e_c$, we will switch to the control output generated by policy B presuming the simulation model is not accurate at this location. And $e_c$ is selected based on policy A's average goal-reaching error: 0.015m. Currently, policy C relies on feedback to verify $e_c$ and operates in a closed-loop configuration.

\subsection{Conclusion}

In this section, we have demonstrated the methods that incorporate neural networks and Gaussian process regression to obtain fast, accurate, and data-efficient control of the continuum robot with soft spine. With advancement of the neural network techniques and a combination of traditional machine learning techniques, controlling such complex and soft robots like malleable robots will become fast and accurate with less effort to perform extensive real-world data calibration. The method we presented is fast, data-efficient and accurate.

\addcontentsline{toc}{section}{References}
\renewcommand\bibname{{References}}
\bibliography{References}
\bibliographystyle{plain}
\end{document}